\documentclass{article}

\PassOptionsToPackage{numbers, compress}{natbib}

    \usepackage[preprint]{neurips_2025}

\usepackage[utf8]{inputenc}
\usepackage[T1]{fontenc}
\usepackage{times}

\usepackage[table,xcdraw]{xcolor}  

\usepackage{hyperref}
\usepackage{url}
\usepackage{microtype}
\usepackage{xspace}
\usepackage{soul}

\usepackage{amssymb}
\usepackage{amsfonts}
\usepackage{amsmath}
\usepackage{amsthm}
\usepackage{nicefrac}

\usepackage{graphicx}
\usepackage{booktabs}
\usepackage{multirow}
\usepackage{adjustbox}
\usepackage{makecell}
\usepackage{colortbl}
\usepackage{wrapfig}
\usepackage{tabularray}   
\usepackage{enumitem}
\usepackage{caption}
\usepackage[font=small]{caption}
\usepackage{subcaption}   
\usepackage{pifont}
\usepackage[ruled,vlined]{algorithm2e}
\usepackage{geometry}

\usepackage{marvosym}

\definecolor{keywordcolor}{RGB}{0,0,150}
\definecolor{commentcolor}{RGB}{0,150,0}
\definecolor{highlightcolor}{RGB}{200,0,0}

\SetKwComment{Comment}{$\triangleright$\ }{}
\SetKwInput{KwInput}{Input} 
\SetKwInput{KwOutput}{Output} 

\theoremstyle{plain}
\newtheorem{theorem}{Theorem}[section]  
\newtheorem{lemma}[theorem]{Lemma}      

\theoremstyle{definition}

\theoremstyle{remark}

\title{
  \raisebox{-0.8ex}{%
    \includegraphics[
      width=0.80cm,
      height=0.80cm,
      keepaspectratio
    ]{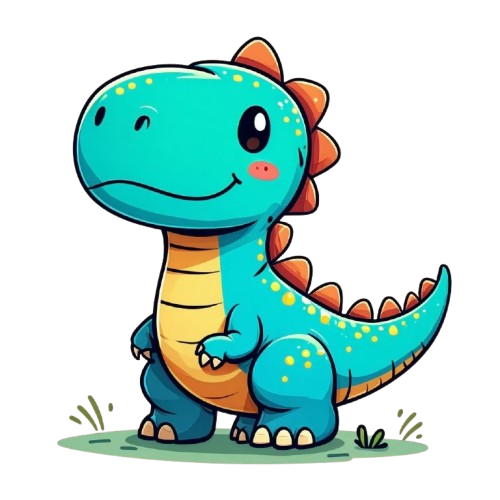}
  }
  T-REX: Mixture-of-Rank-One-Experts with Semantic-aware Intuition for Multi-task Large Language Model Finetuning}

\author{%
Rongyu Zhang$^{1,2,3}$\thanks{Equal contributions; $\textrm{\Letter}$ Corresponding authors.} , Yijiang Liu$^{1}$\footnotemark[1] , Huanrui Yang$^{4}$\footnotemark[1] , Shenli Zheng$^{1}$ \\
\textbf{Chongkang Tan$^{5}$, Dan Wang$^{2}$, Yuan Du$^{1}$, Li Du$^{1,\textrm{\Letter}}$, Shanghang Zhang$^{3, \textrm{\Letter}}$}\\
$^{1}$ Nanjing University 
$^{2}$ PolyU
$^{3}$ Peking University
$^{4}$ University of Arizona
$^{5}$ Ant Group \\
}

\begin{document}

\maketitle

\begin{abstract}
Large language models (LLMs) encounter significant adaptation challenges in diverse multitask finetuning. 
Mixture-of-experts (MoE) provides a promising solution with a dynamic architecture, enabling effective task decoupling. However, scaling up the number of MoE experts incurs substantial parameter and computational overheads and suffers from limited performance gain due to naive routing mechanisms. 
In this paper, we design a novel framework, mix\underline{\textbf{T}}ure\underline{\textbf{-}}of-\underline{\textbf{R}}ank-on\underline{\textbf{E}}-e\underline{\textbf{X}}perts (\texttt{T-REX}), which leverages the combination of ultra-low rank experts to construct LoRA weights on pretrained LLMs. The rank-1 experts enable a mix-and-match mechanism to quadratically expand the vector subspace of experts with linear parameter overheads, achieving approximate error reduction with optimal efficiency.
In addition, T-REX offers implicit guidance to the router, leveraging the inherent semantic clustering of training embeddings as prior knowledge, enabling optimized feature allocation across experts for a smoother convergence.
Extensive theoretical and empirical results demonstrate that T-REX achieves superior efficiency and generalizability across diverse tasks. Compared with other LoRA-based methods, T-REX achieves up to 1.78\% mean accuracy improvement with around 30\%-40\% less trainable parameters across 14 public datasets. \href{https://github.com/RoyZry98/T-REX-Pytorch}{Code} is available.
\end{abstract}

\section{Introduction}
Large language models (LLMs)~\cite{Touvron2023Llama2O,Jiang2023Mistral7,Mesnard2024GemmaOM} exhibit profound capabilities across various domains, excelling in specific downstream tasks including content generation, enhancing interactive entertainment, and inspiring artistic endeavors. However, despite the transformative impact of LLMs on natural language processing (NLP), as evidenced by their mastery of unsupervised pretraining and subsequent supervised fine-tuning~\cite{alpaca, Hu2021LoRALA}, the intricate settings of complex multitask learning (MTL) environments present a unique challenge. The diversity of tasks in these environments tests the adaptability of LLMs~\cite{zhang2023llama,sung2022vl}, pushing the boundaries of their application and capability.

Mixture-of-Experts (MoE)~\cite{masoudnia2014mixture,moe,stablemoe,ori_moe1} has recently gained attention as a practical approach for multi-task finetuning, leveraging its dynamic architecture to adapt the generally pretrained LLMs to diverse downstream tasks efficiently. Like humans, who demonstrate superior multitasking abilities by selectively activating specific regions of the brain based on explicit cues, MoE dynamically engages and combines different experts for each particular task. However, applying MoE to LLM multi-task learning raises several unresolved challenges. \textbf{\ding{192} Significant parameter and computational overheads}: Unlike single-task MoE, where only a few experts suffice, multitask MoE requires scaling the number of experts with the number of tasks, leading to significant costs in training and storage. While prior works~\cite{zadouri2023pushing,zhu2023sira} have explored parameter-efficient finetuning (PEFT) experts using Low-rank adapter (LoRA)~\cite{Hu2021LoRALA,huang2023lorahub}, the parameter overhead still scales linearly with the number of experts. \ding{193} \textbf{Unfair router task allocation}: Traditional MoE routers, often based on a linear layer with softmax, struggle with fine-grained task differentiation, resulting in suboptimal expert allocation. Recent approaches~\cite{luo2023mowe,zhang2024efficient} have introduced uncertainty-aware mechanisms~\cite{lakshminarayanan2017simple,zhang2023unimodal,liu2022multi} and explicit feature guidance to enhance routing, but their effectiveness in MTL remains unproven.

To address these challenges, we propose a novel paradigm, mix\underline{\textbf{T}}ure\underline{\textbf{-}}of-\underline{\textbf{R}}ank-on\underline{\textbf{E}}-e\underline{\textbf{X}}perts (\texttt{T-REX}), which leverages \ding{202} ultra-lightweight rank-1 experts with a mix-and-match mechanism. 
Specifically, T-REX allows multiple rank-1 vectors $a_i\in\mathbb{R}^{m\times 1}$ and $b_j\in\mathbb{R}^{n\times 1}$ to pair freely into rank-1 LoRA experts $\Delta W_{ij} = a_i b_j^T$. This mix-and-match mechanism enables T-REX to construct more LoRA experts to cover a higher-dimensional subspace in the weight matrix space with a sub-linear growth of additional parameters. 
This mix-and-match strategy also facilitates more effective interaction and cross-fusion of shared feature representations across tasks, ultimately boosting the model's generalization capability. Mathematically, T-REX is proven to increase representational capacity over other LoRA-based methods, therefore achieving higher accuracy empirically across tasks.

In addition, we propose \ding{203} a semantic-aware router inspired by the brain's ability for intuitive cognition, enabling T-REX to better allocate experts for complex multitask scenarios. Naively, the router can utilize the human-annotated task categorization as a condition proxy as \textit{explicit intuition}. 
However, this empirically leads to 2.50\% and 3.77\% accuracy drops for LLaMA-2~\cite{Touvron2023Llama2O} and Gemma~\cite{Mesnard2024GemmaOM} on OpenCompass\footnote{\url{https://github.com/open-compass/opencompass}}, respectively. 
This highlights that the human-labeled training tasks are coarse-grained and each requires a mixture of diverse skills, whose ambiguity contradicts the expert's need to identify and learn task-specific capabilities. To address this, we redefine expert tasks using the clustering of training embeddings. We assign inputs to experts based on the similarity between input embeddings and the corresponding task centroids. Specifically, T-REX leverages the correlation between the input and these centroids as an \textit{implicit intuition}, which conditions the router’s decision-making process and guides all experts to converge smoothly and achieve generalizability.

Both theoretical derivations and extensive experiments demonstrate the superiority of our proposed T-REX against state-of-the-art baselines, notably improving both the performance and the computational efficiency, delivering an increase in overall accuracy by up to 1.78\% with around 30\%-40\% less parameter overhead than other LoRA-based methods. The contributions of T-REX include:
\begin{itemize}
    \item We devise a novel LoRA MoE paradigm that elaborately leverages ultra-lightweight rank-1 experts with a mix-and-match mechanism for efficient multi-task finetuning. 
     \item Drawing on the concept of implicit intuition akin to that of a human brain, we provide conditional guidance for the router by leveraging the resemblance between the input task instance and predefined embedding clusters.  
    \item We provide a theoretical analysis showing that T-REX achieves expert linear combination subspace expansion, enabling approximate error reduction, while guiding experts to converge smoothly, ultimately enhancing overall model performance.

\end{itemize}
\vspace{-1em}

\section{Related Work}
\vspace{-0.5em}
\paragraph{Multi-task learning.}
MTL aims to enhance a model's generalization across multiple tasks by sharing insights. The shared-bottom model~\cite{caruana1997multitask} uses hard-shared parameters but can suffer from negative transfer. To address this, studies~\cite{kendall2018multi,chen2018gradnorm} propose adaptive loss weighting to balance multi-task losses. Models like Cross-stitch~\cite{misra2016cross} and Sluice~\cite{ruder2019latent} networks dynamically blend task-specific representations but use static task weights. Our method utilizes a MoE to achieve precise task allocation by assigning input to a combination of specialized experts.

\vspace{-0.5em}
\paragraph{Mixture of Experts.}
MoE was initially introduced by ~\cite{ori_moe1,ori_moe2} to process different samples with independent modules. For example, ~\cite{moe} employs MoE in large-scale LSTM-based~\cite{lstm} models. GShard~\cite{gshard} and Switch Transformer~\cite{switch} firstly introduce MoE in Transformer, and largely scale up model size with top-1/2 routing strategies.~\cite{hash,stablemoe} stabilize the training process with fixed routing strategies. ~\cite{ec} refines the routing strategy, which allows tokens to be assigned to varying numbers of experts. ST-MoE~\cite{st_moe} optimizes the training instability and fine-tuning complexity in MoE models.
~\cite{zadouri2023pushing,zhu2023sira} propose parameter-efficient adaptors as experts to decrease MoE size.~\cite{puigcerver2023sparse} proposes a fully-differentiable sparse MoE to address the training instability and token dropping problems. Our method enhances routing with multi-task data insights and employs a rank-1 expert for efficiency.

\paragraph{Parameter-efficient finetuning.}
PEFT~\cite{fu2023effectiveness,ding2023parameter,zaken2021bitfit} adjusts minimal parameters to reduce storage demands. PEFT methods include specification-based~\cite{zaken2021bitfit}, reparameterization-based~\cite{ding2021repvgg,zhang2023repcam}, and addition-based approaches.
The most prevailing addition-based PEFT, like Adapter-Tuning~\cite{zhang2023llama}, Prefix Tuning~\cite{li2021prefix}, and Prompt Tuning~\cite{lester2021power}, adds new modules to the base model. Recent studies have combined low-rank adapters (LoRA)~\cite{Hu2021LoRALA} with MoE to enhance flexibility in PEFT. However, existing LoRA-MoE methods~\cite{zadouri2023pushing,zhu2023sira} face challenges in effectively balancing the parameter efficiency and the scalability to more experts. To address this, our proposed T-REX incorporates rank-1 mix-and-match experts to scale up LoRA MoE with minimal overhead.

\section{Methodology}
\label{sec:method}

\subsection{Preliminary}
\paragraph{Problem formulation.}
In multitask learning with $K$ distinct tasks, each task $k$ is defined over an input space $\mathcal{X}_k \subset \mathbb{R}^m$, an output space $\mathcal{Y}_k$, and a distinct objective such as classification, generation, or retrieval. Given a pretrained weight matrix $\mathbf{W}_0 \in \mathbb{R}^{m \times n}$ representing the parameters of a LLM, the goal of the multitask fine-tuning is to adapt $\mathbf{W}_0$ for each task efficiently while avoiding catastrophic forgetting. For an input $\mathbf{x} \in \mathcal{X}_k$ from task $k$, the model’s output $\mathbf{z} \in \mathcal{Y}_k$ is computed as $\mathbf{z} = f(\mathbf{W}_0, \mathbf{x})$, where $f(\cdot)$ denotes the forward pass to generalize across heterogeneous domains.

\paragraph{LoRA-MoE.}
LoRA-MoE~\cite{zadouri2023pushing} introduces $N$ experts added onto $\mathbf{W}_0$, and a router dynamically selects and combines these experts based on the input. As illustrated in Fig~\ref{fig:trex} (b), the $i$-th expert is parameterized as: $f_i(\mathbf{x}) = \Delta \mathbf{W}_i \mathbf{x}$, where $\Delta \mathbf{W}_i$ is a low-rank matrix achieved via parameter-efficient finetuning methods like LoRA. The overall layer output is computed as:
\begin{equation}
    \mathbf{z} = \mathbf{W}_0 \mathbf{x} + \underbrace{\sum_{i=1}^N \mathbf{G}(\mathbf{x}) \Delta \mathbf{W}_i \mathbf{x}}_{\text{Expert Contribution}},
    \label{eq:moe_formal_simple}
\end{equation}
where $\mathbf{G}(\mathbf{x})$ represents the routing weight for the $i$-th expert, dynamically computed for each input $\mathbf{x}$.

\subsection{Mix-and-Match for rank-1 experts}
\label{sec:rank1-experts}

\begin{figure}[t]
\centering
\includegraphics[width=\linewidth]{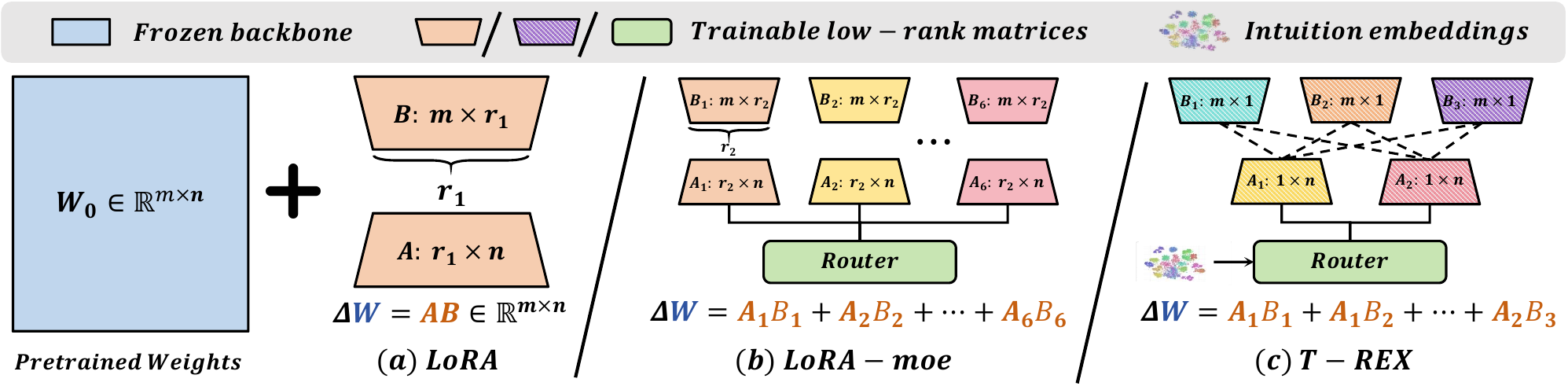}
\caption{Illustration of the trainable parameters in (a) Vanilla LoRA, (b) LoRA-MoE, and (c) our proposed T-REX. $r_{1},r_{2}\ll min\{m,n\}$. 6 experts are demonstrated for both LoRA-MoE and T-REX. }
\label{fig:trex}
\end{figure}

To tackle the scalability challenges of MoE, we introduce Rank-1 Experts, designed to maximize parameter efficiency and enhance the flexibility of MoE. 
Specifically, we start with a decoupled row-column subspace decomposition of the LoRA matrix, as shown in Fig~\ref{fig:trex} (c). Formally, for a pretrained weight matrix $\mathbf{W}_0 \in \mathbb{R}^{m \times n}$, the $i$-th rank-1 LoRA expert is defined as:
\begin{equation}
\Delta \mathbf{W}_i =\mathbf{a}_i \mathbf{b}_i^\top, \quad \text{where} \quad \mathbf{a}_i \in \mathbb{R}^m, \ \mathbf{b}_i \in \mathbb{R}^n.
\label{eq:rank1_expert}
\end{equation}
$\{\mathbf{a}_i\}$ and $\{\mathbf{b}_i\}$ represent trainable base vectors spanning the row and column subspaces, respectively. With $N$ such experts, the layer output is computed as:
\begin{equation}
\mathbf{z} = \mathbf{W}_0\mathbf{x} + \mathbf{A} \cdot \mathrm{diag}(\mathbf{G}(\mathbf{x})) \cdot \mathbf{B}^\top \mathbf{x},
\label{eq:moe_base}
\end{equation}
where $\mathbf{A} = [\mathbf{a}_1 | \cdots | \mathbf{a}_N]$ and $\mathbf{B} = [\mathbf{b}_1 | \cdots | \mathbf{b}_N]$ form the factorized LoRA weights.

The above rank-1 MoE serves as the most compact design of the LoRA-MoE. However, the number of experts in such design is still limited by the rank of LoRA weights $\mathbf{A}$ and $\mathbf{B}$, where at most $N$ experts can be established given rank-$N$ LoRA weights.
To enhance flexibility and reduce parameter overheads in scaling up the number of experts, we decouple the one-to-one correspondence between the row and column subspaces represented by $\mathbf{a}_i$ and $\mathbf{b}_i$ in conventional LoRA. Specifically, any $\mathbf{a}_i$ can be paired with any $\mathbf{b}_j$, forming a mix-and-match expert $f_{ij}(\mathbf{x}) = \mathbf{a}_i \mathbf{b}_j^\top \mathbf{x}$. This mechanism expands the expert pool from a linear scaling of the rank to a quadratic scaling by enabling cross-combinations. The overall output is computed as:
\begin{equation}
\mathbf{z} = \mathbf{W}_0 \mathbf{x} + \sum_{i=1}^I \sum_{j=1}^J G_{ij}(\mathbf{x}) (\mathbf{a}_i \mathbf{b}_j^\top) \mathbf{x} = \mathbf{W}_0\mathbf{x} + \mathbf{A} \cdot \mathbf{G}(\mathbf{x}) \cdot \mathbf{B}^\top \mathbf{x},
\label{eq:moe_mix}
\end{equation}
where $I=\text{Rank}(\mathbf{A})$, $J=\text{Rank}(\mathbf{B})$, and the routing matrix $\mathbf{G}(\mathbf{x}) \in \mathbb{R}^{I \times J}$ no longer required to be square or diagonal, dynamically activates these cross-combined experts. Furthermore, the mix-and-match mechanism can be expressed as a Kronecker product expansion:
\begin{equation}
\sum_{i,j=1}^N G_{ij}(\mathbf{x}) (\mathbf{a}_i \otimes \mathbf{b}_j) = (\mathbf{A} \otimes \mathbf{B})\mathrm{vec}(\mathbf{G}),
\end{equation}
where $\mathbf{A} \otimes \mathbf{B}$ represents the Kronecker product of the row and column subspaces, $\mathrm{vec}(\mathbf{G})$ flattens the routing matrix, and $N=I\times J$ is the total number of experts. This design achieves quadratic scaling of expressive power with only linearly increased additional parameters, fundamentally differing from the capacity-parameter tradeoff of standard LoRA. The pseudo-code is shown in Algorithm~\ref{alg:T-REX}.

\subsection{Semantic-aware router with intuition clustering}
\label{sec:routing}
Naive MoE lacks a meaningful correlation between the expert routing and the task, leading to either under-trained or redundant experts. 
In T-REX, we propose to implicitly assign a specific ``task'' for each expert to specialize in, therefore effectively covering the MTL space. However, we observe that the human-defined training task in the MTL process often leads to ambiguity in the data distribution and the model capability required. 
For example, we visualize token embeddings of 20 datasets across five human-defined categories\footnote{Details are shown in the Appendix~\ref{ap:motivation}.}. We observe that embeddings of the same task category distribute throughout the space, whereas some embeddings from diverse tasks appear in close clusters. This demonstrates that predefined training tasks are coarse-grained and involve diverse skills. 
To this end, we propose to design the ``task'' of each expert based on the semantic clusters of the training set embeddings. Each expert will be encouraged to better process the tokens within a semantic cluster, while the token will be routed to the experts based on its similarity with the cluster centroids.

Formally, suppose we have $N$ experts in the T-REX. We start by gathering a set of embeddings from the multitask training tokens $\hat{\mathbf{e}} = \boldsymbol{E}(\hat{\mathbf{x}}) \in \mathbb{R}^d$ using an Ada-embedding model. These embeddings are then grouped into $N$ semantic clusters, corresponding to $N$ experts, using $k$-means clustering:
\begin{equation}
\mathcal{C}_i = \{\hat{\mathbf{e}}_j : ||\hat{\mathbf{e}}_j - \boldsymbol{\mu}_i||^2 \leq ||\hat{\mathbf{e}}_j - \boldsymbol{\mu}_r||^2,\ \forall r \in [1, N]\},
\label{eq:clustering}
\end{equation}
where $\boldsymbol{\mu}_i \in \mathbb{R}^d$ is the centroid of the $i^{th}$ cluster. During finetuning, we compute \textit{intuition scores} $\mathbf{I}(\mathbf{x}) \in \mathbb{R}^N$ for an input $\mathbf{x}$ using cosine similarity between its embedding $e$ and the cluster centroids. For each cluster centroid $\boldsymbol{\mu}_i$, the intuition score $\mathbf{I}_i$ measures the similarity:
\begin{equation}
\label{equ:intuition}
\mathbf{I}_i(\boldsymbol{x}) = \cos(\angle(\boldsymbol{e}, \boldsymbol{\mu}_i)) = \frac{\boldsymbol{e}^\top \boldsymbol{\mu}_i}{||\boldsymbol{e}||\cdot||\boldsymbol{\mu}_i||}.
\end{equation}
The intuition vector $\mathbf{I}(\mathbf{x}) \in \mathbb{R}^N$ gathering all intuition scores is then fused with the base routing weights $\mathbf{G}(\mathbf{x}) \in \mathbb{R}^N$ through element-wise addition:
\begin{equation}
\widetilde{\mathbf{G}}(\mathbf{x}) = \mathbf{G}(\mathbf{x}) \oplus \mathbf{I}(\mathbf{x}),
\label{eq:fusion}
\end{equation}
where $\oplus$ denotes element-wise addition. This enhanced routing signal is used to construct the final expert combination, enabling a more semantically informed selection of experts.

\begin{algorithm}[t]
\DontPrintSemicolon
\SetKwInput{KwInit}{Initialize}
\SetKwInput{KwReturn}{Return}
\SetKwInOut{KwRequire}{Require}
\SetKwInOut{KwEnsure}{Ensure}

\KwInput{Pre-trained weight matrix $\mathbf{W}_0 \in \mathbb{R}^{m \times n}$, inputs $X$, LoRA weights ranks $I, J$, task centroids $\{\boldsymbol{\mu}_k\}_{k=1}^N, N=I\times J$, embedding model $\boldsymbol{E}$, maximum iterations $\texttt{MaxIter}$.}
\KwOutput{Adapted model weights $\mathbf{W}_{\text{adapted}}$.}

\KwInit{
    $\{\mathbf{a}_i, \mathbf{b}_j\}$ for $i = 1\ldots I, j = 1 \ldots J$,
    routing weight matrix $\mathbf{G}(\cdot) \in \mathbb{R}^{I \times J}$. 
}

\For{$\texttt{Iter} = 1$ to $\texttt{MaxIter}$}{
    \Comment*[l]{\textcolor{keywordcolor}{Step 1: Compute Routing Weights with Intuition}}
    \ForEach{$\mathbf{x} \in X$}{
        Base routing: $\mathbf{g}_{\text{base}} = \texttt{softmax}(\mathbf{G}(\mathbf{x}))$.\;
        Intuition scores: $\mathbf{I}(\mathbf{x}) = [\cos(\boldsymbol{E}(\mathbf{x}), \boldsymbol{\mu}_k)]_{k=1}^N$.\;
        Fuse routing: $\mathbf{\widetilde{G}}(\mathbf{x}) = \texttt{softmax}(\mathbf{g}_{\text{base}} \oplus \mathbf{I}(\mathbf{x}))$. \Comment*[r]{\textcolor{commentcolor}{Enhanced routing}}
    }

    \Comment*[l]{\textcolor{highlightcolor}{Step 2: Compute Mix-and-Match with Rank-1 Experts}}
    \begin{equation}
    \Delta \mathbf{W}_{\text{combined}}(\mathbf{x}) = \sum_{i=1}^I \sum_{j=1}^J \mathbf{\widetilde{G}}_{ij}(\mathbf{x}) \mathbf{a}_i \mathbf{b}_j^\top
    \end{equation}

}

\KwReturn{$\mathbf{W}_{\text{adapted}} = \mathbf{W}_0 + \Delta \mathbf{W}_{\text{combined}}$.}

\caption{\textbf{T-REX}: Mixture of Rank-1 Experts with Semantic-Aware Routing}
\label{alg:T-REX}
\end{algorithm}

\section{Theoretical analysis of T-REX}
\label{sec:theory}

\subsection{Rank-1 experts with subspace expansion}
\label{sec:rank1-theory}
\begin{lemma}[Subspace Expansion with Mix-and-Match]
The adaptation matrix $\Delta W = \mathbf{A}\mathbf{G}\mathbf{B}^\top$ generated by the Mix-and-Match mechanism spans a subspace whose dimensionality grows with a speed of $\mathcal{O}(IJ)$ as LoRA weight ranks $I$ and $J$ increase. Specifically, the vectorized form of $\Delta W$, $\mathrm{vec}(\Delta W)$, spans a space with dimensionality up to:
\begin{equation}
\mathrm{dim}(\mathrm{span}(\mathrm{vec}(\Delta W))) = I\times J,
\end{equation}
while the rank of the matrix $\Delta W$ itself is bounded by $\mathrm{rank}(\Delta W) \leq \min{(I,J)}$.
\end{lemma}

\begin{proof}
Under the Mix-and-Match mechanism, any $\mathbf{a}_i$ can pair with any $\mathbf{b}_j$, forming experts $f_{ij}(\mathbf{x}) = \mathbf{a}_i \mathbf{b}_j^\top \mathbf{x}$. Thus, the vectorized form of $\Delta W$ can be expressed as:
\begin{equation}
\mathrm{vec}(\Delta W) = (\mathbf{A} \otimes \mathbf{B}) \mathrm{vec}(\mathbf{G}),
\end{equation}
where $\mathbf{A} \in \mathbb{R}^{m \times I}$, $\mathbf{B} \in \mathbb{R}^{n \times J}$ are the Rank-1 expert basis matrices, $\mathbf{G} \in \mathbb{R}^{I \times J}$ is the dynamic routing matrix, and $\otimes$ denotes the Kronecker product. The Kronecker product $\mathbf{A} \otimes \mathbf{B}$ spans a subspace with dimensionality:
\begin{equation}
\mathrm{dim}(\mathrm{span}(\mathbf{A} \otimes \mathbf{B})) = \mathrm{dim}(\mathcal{S}_a \otimes \mathcal{S}_b) = I\times J,
\end{equation}
which significantly expands the potential representation capability of $\Delta W$. 
\end{proof}

\begin{theorem}[Approximation Error Bound with Rank-1 Experts]
For a target adaptation matrix $\Delta W^\star$, T-REX decomposes the matrix into in-subspace and residual components:
\begin{equation}
\Delta W^\star = \Delta \mathbf{W}_\mathrm{in} + \Delta \mathbf{W}_\mathrm{out}, \quad \Delta \mathbf{W}_\mathrm{in} \in \mathcal{S}_a \otimes \mathcal{S}_b, \quad \Delta \mathbf{W}_\mathrm{out} \perp \mathcal{S}_a \otimes \mathcal{S}_b.
\end{equation}
The Frobenius norm approximation error satisfies:
\begin{equation}
\|\Delta W - \Delta W^\star\|_F^2 \leq \|\Delta \mathbf{W}_\mathrm{out}\|_F^2 + C \cdot \frac{1}{IJ} \|\Delta W^\star\|_F^2 + \mathcal{O}\left(\frac{1}{\tau}\right),
\end{equation}
where:
- $\|\Delta \mathbf{W}_\mathrm{out}\|_F^2$ quantifies the residual error outside the subspace $\mathcal{S}_a \otimes \mathcal{S}_b$;
- The second term reflects the alignment error of $\Delta W^\star$ with the subspace, which decreases as $IJ$ increases;
- The third term $\mathcal{O}(1/\tau)$ accounts for the effect of the softmax temperature $\tau$ in the dynamic routing matrix $\mathbf{G}(\mathbf{x})$, which sharpens the subspace activation.
\end{theorem}

\begin{proof}
By the orthogonal projection theorem, the target matrix $\Delta W^\star$ can be decomposed into in-subspace $\Delta \mathbf{W}_\mathrm{in}$ and residual $\Delta \mathbf{W}_\mathrm{out}$ components. The in-subspace component is represented by $(\mathbf{A} \otimes \mathbf{B}) \mathrm{vec}(\mathbf{G})$. Thus, the approximation error can be written as:
\begin{equation}
\|\Delta W - \Delta W^\star\|_F^2 = \|\Delta \mathbf{W}_\mathrm{out}\|_F^2 + \|\Delta \mathbf{W}_\mathrm{in} - \Delta W^\star\|_F^2.
\end{equation}
For the subspace alignment error $\|\Delta \mathbf{W}_\mathrm{in} - \Delta W^\star\|_F^2$, the error decreases with the subspace dimensionality $IJ$. Specifically, the error term satisfies:
\begin{equation}
\|\Delta \mathbf{W}_\mathrm{in} - \Delta W^\star\|_F^2 \propto \frac{1}{IJ} \|\Delta W^\star\|_F^2.
\end{equation}
Additionally, the routing matrix $\mathbf{G}(\mathbf{x})$'s softmax temperature $\tau$ controls the activation sharpness of virtual experts. As $\tau$ increases, $\mathbf{G}(\mathbf{x})$ becomes more selective, reducing the error fluctuation, with a contribution approximated as $\mathcal{O}(1/\tau)$. A detailed derivation is provided in Appendix \ref{ap:theory}.
\end{proof}

\begin{figure}[t]
\centering
\includegraphics[width=\linewidth]{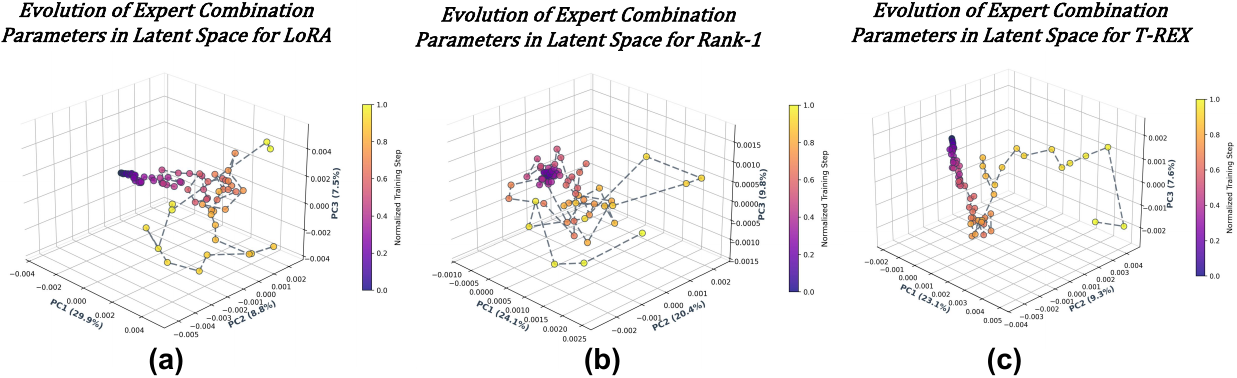}
\caption{3D trajectory of router weights for (a) Traditional LoRA, (b) Rank-1 experts with Mix-and-Match, and (c) T-REX with intuition in the MoE training process. Intuition helps to enable a smoother convergence.}
\label{fig:manifold}
\end{figure}
\vspace{-1em}

\subsection{Intuition leads to low-loss region}
\label{sec:intuition_analysis}

The previous section proves the ability of mix-and-match rank-1 experts to optimally approximate the adaption matrix of a specific task. In this section, we discuss how the intuition vector proposed in Eq.~\ref{equ:intuition} guide the combination of task-specific experts to fulfill diverse tasks.

\begin{theorem}[Expert Combination with Intuition Guidance]
For a multitask learning task objective defined on an input space $\mathcal{X} \subset \mathbb{R}^{1\times m}$ as
\begin{equation}
\label{equ:MTL_objective}
    \mathcal{L}(\Delta W,x) = \sum_{i=1}^N ||x\Delta W - \mu_i \Delta W_i^\star||^2,
\end{equation}
where $\mu_1,\cdots \mu_N$ forms the orthonormal basis of the input subspace $\mathcal{X}$ and $\Delta W_1^\star,\cdots, \Delta W_N^\star$ are the target adaption matrices for the task-specific experts, the intuition vector $\mathbf{I}_i(\boldsymbol{x}) = \cos(\angle(\boldsymbol{x}, \boldsymbol{\mu}_i))$ is proportional to the coordinate vector of the optimal $\Delta W$ under the basis of $\Delta W_1^\star,\cdots, \Delta W_N^\star$.
\end{theorem}
\begin{proof}
Eq.~\ref{equ:MTL_objective} can be minimized by solving the gradient equation
\begin{equation}
    \frac{\partial\mathcal{L}}{\partial\Delta W} = \sum_{i=1}^N x^T(x\Delta W - \mu_i \Delta W_i^\star) = 0; \text{solved when}\ \Delta W = \frac{\sum x^T\mu_i \Delta W_i^\star}{N x^Tx}.
\end{equation}
Therefore the coordinate vector of the optimal $\Delta W$ under the basis of $\Delta W_1^\star,\cdots, \Delta W_N^\star$ is proportional to $x^T \mu_i$.
Since $\mu_1,\cdots \mu_N$ forms the orthonormal basis of $\mathcal{X}$, the input $x$ can be represented as $x = \sum_{i=1}^N \alpha_i \mu_i$, where $\alpha_i = \cos(\angle(\boldsymbol{x}, \boldsymbol{\mu}_i)) := \mathbf{I}_i(\boldsymbol{x})$ by definition, we have $x^T\mu_i = \mathbf{I}_i(\boldsymbol{x})$. 
\end{proof}
In the case of T-REX, though much more complicated than the task in the theorem, we show that that intuition vector serves as a good initialization and guidance for the learnable router to decompose the optimal adaptation matrix of each input to the basis of mix-and-match rank-1 experts. This is demonstrated by the smoother router convergence trajectory evidenced in Fig.~\ref{fig:manifold}.

\section{Experiment}

\subsection{Implementation Details}
\label{sec:exp-baseline}
We leverage an extensive collection of open-source LLM, including LLaMA-3~\cite{grattafiori2024llama}, LLaMA-2~\cite{Touvron2023Llama2O}, Mistral~\cite{Jiang2023Mistral7}, Yi~\cite{young2024yi}, 
Phi~\cite{javaheripi2023phi}, Gemma~\cite{Mesnard2024GemmaOM}, and Tiny-LLaMA~\cite{Zhang2024TinyLlamaAO}. The MoE module is implemented in the transformer blocks, explicitly targeting the query, key, value, and feed-forward network components, aligning with conventional LoRA standards. To generate instance embeddings, we utilize nomic-embed-text-v1~\cite{nussbaum2024nomic}, the most downloaded embedding model on Hugging Face.

Our PEFT procedures are performed with a 64 batch size, a 5e-5 learning rate adjusted with a cosine learning rate scheduler and Adam optimizer, spanning three epochs.
All experiments are conducted on NVIDIA A100 GPUs, with the GPU hours varying between 30 and 300.

\vspace{-0.5em}
\subsection{Multitask Datasets and Baselines}
\label{sec:exp-dataset}
\textbf{Datasets.} We curate a comprehensive multitask dataset by aggregating twenty datasets from diverse sources, spanning five empirically determined dimensions~\cite{2023opencompass}: reasoning, examination, language, understanding, and knowledge. The reasoning category includes datasets such as ANLI~\cite{nie2019adversarial}, ReCoRD~\cite{wang2019superglue}, and HellaSwag~\cite{zellers2019hellaswag}; for examination, we feature MMLU~\cite{hendryckstest2021} and ARC~\cite{allenai:arc}; WiC~\cite{wang2019superglue} and WinoGrande~\cite{ai2:winogrande} represent language; OpenBookQA~\cite{OpenBookQA2018} and MultiRC~\cite{wang2019superglue} embody understanding; CommonSenseQA~\cite{talmor-etal-2019-commonsenseqa} and BoolQ~\cite{wang2019superglue} epitomize knowledge. Additionally, Alpaca~\cite{alpaca}, which encompasses mixed dimensions, is also included in our training set.

\noindent\textbf{Baselines.} The comparative methodologies include conventional LoRA~\cite{Hu2021LoRALA} with the rank values set to 32, marked as LoRA-32. Additionally, we involve two state-of-the-art LoRA-based MoE methods, including MoLoRA~\cite{zadouri2023pushing}, which introduces lightweight experts and is capable of generalizing to new tasks independently of prior knowledge, and SiRA\cite{zhu2023sira} that enhances LoRA by integrating sparse MoE with capacity-limited top-k expert routing and introduces a novel expert dropout method to mitigate overfitting. 
The experts in SiRA and MoLoRA are set as 8 $\times$ LoRA-4, aligning the learnable parameters of LoRA-32, while our T-REX adopts a total of 32 Rank-1 experts via the mix-and-match of rank-4 and rank-8 LoRA weights in the main results.

\setlength\tabcolsep{3pt}
\begin{table*}[t]
\centering
\caption{Performance evaluation across 14 datasets on 7 LLM backbones compared with three baselines. MUL. indicates MULTIRC and BOO. indicates BOOLQ. \#ATP stands for additional trainable parameters.}
\label{tab:my-table}
\resizebox{1\textwidth}{!}{%
\begin{tabular}{c|cccccccccccccc|cc}
\toprule
 Method & MUL. & MMLU & BOO. & WIC & WG & WSC & ANLI & PIQA & SIQA & RTE & COPA & OBQA & CSQA & HS & \#ATP \textcolor{red}{\textbf{\textcolor{red}{\textbf{$\downarrow$}}}} & AVG \textcolor{rgb:red,0.0;green,0.8;blue,0.2}{\textbf{\textcolor{rgb:red,0.0;green,0.8;blue,0.2}{\textbf{$\uparrow$}}}} \\ 
 \midrule
 \midrule
\rowcolor{gray!10}
\multicolumn{17}{c}{\textit{\textbf{LLaMA-2 13B}}} \\
\midrule
\midrule
  LoRA & 90.31 & 60.48 & 90.13 & \textbf{75.47} & \textbf{87.66} & 64.42 & 76.10 & 88.70 & 82.40 & 90.71 & \textbf{100.00} & 84.92 & 84.80 & 96.49 & +0.95\% & 83.76 \\
  MoLoRA & 90.35 & \textbf{62.30} & 89.73 & 74.84 & 86.50 & 66.35 & \textbf{76.90} & 87.83 & 82.60 & 90.00 & 98.00 & 85.71 & 84.64 & 96.36 & +1.06\% & 83.73 \\
  SiRA & 90.20 & 60.94 & 90.34 & 74.61 & 87.19 & 67.31 & 76.30 & 88.79 & 82.96 & 90.61 & \textbf{100.00} & 84.20 & 84.44 & 95.44 & +1.06\% & 83.81 \\
 \cellcolor{green!10}T-REX & \cellcolor{green!10}\textbf{90.98} & \cellcolor{green!10}59.33 & \cellcolor{green!10}\textbf{91.24} & \cellcolor{green!10}73.26 & \cellcolor{green!10}86.71 & \cellcolor{green!10}\textbf{69.27} & \cellcolor{green!10}76.40 & \cellcolor{green!10}\textbf{88.98} & \cellcolor{green!10}\textbf{83.65} & \cellcolor{green!10}\textbf{93.78} & \cellcolor{green!10}98.00 & \cellcolor{green!10}\textbf{86.40} & \cellcolor{green!10}\textbf{86.42} & \cellcolor{green!10}\textbf{96.67} & \cellcolor{green!10}\textcolor{red}{\textbf{+0.62\%}} & \cellcolor{green!10}\textcolor{rgb:red,0.13;green,0.55;blue,0.13}{\textbf{84.37}} \\
  \midrule
 \midrule
\rowcolor{gray!10}
\multicolumn{17}{c}{\textit{\textbf{LLaMA-3 8B}}} \\
\midrule
\midrule
  LoRA & 90.00 & \textbf{61.87} & 90.03 & 74.39 & 84.54 & \textbf{65.25} & 75.70 & 89.36 & 83.06 & 89.95 & \textbf{99.00} & 88.40 & 84.38 & 95.10 & +1.03\% & 83.80 \\
  MoLoRA & 89.91 & 61.01 & 90.00 & 74.14 & 83.98 & 58.65 & 72.60 & 89.88 & 82.96 & 88.09 & 97.00 & 89.40 & 83.70 & \textbf{95.49} & +1.16\% & 82.63 \\
  SiRA & 90.18 & 61.07 & \textbf{90.52} & \textbf{74.76} & 85.56 & 63.46 & 75.90 & \textbf{90.26} & \textbf{83.88} & 88.81 & \textbf{99.00} & 87.80 & \textbf{84.93} & 95.52 & +1.16\% & 83.69 \\
 \cellcolor{green!10}T-REX & \cellcolor{green!10}\textbf{90.35} & \cellcolor{green!10}61.33 & \cellcolor{green!10}90.43 & \cellcolor{green!10}74.92 & \cellcolor{green!10}\textbf{86.19} & \cellcolor{green!10}65.38 & \cellcolor{green!10}\textbf{78.10} & \cellcolor{green!10}89.45 & \cellcolor{green!10}83.01 & \cellcolor{green!10}\textbf{90.25} & \cellcolor{green!10}\textbf{99.00} & \cellcolor{green!10}87.80 & \cellcolor{green!10}\textbf{84.93} & \cellcolor{green!10}94.95 & \cellcolor{green!10}\textcolor{red}{\textbf{+0.69\%}} & \cellcolor{green!10}\textcolor{rgb:red,0.13;green,0.55;blue,0.13}{\textbf{84.01}} \\ 
 \midrule
 \midrule
\rowcolor{gray!10}
\multicolumn{17}{c}{\textit{\textbf{LLaMA-2 7B}}} \\
\midrule
\midrule
  LoRA & 88.00 & 52.71 & \textbf{88.56} & 70.69 & 80.51 & 57.69 & \textbf{73.80} & 84.93 & 81.47 & 86.28 & 95.00 & 82.20 & 83.29 & 93.50 & +1.17\% & 79.90 \\
  MoLoRA & \textbf{88.90} & \textbf{54.02} & 87.80 & 70.38 & 78.69 & 53.85 & 71.60 & 84.82 & 82.60 & 87.36 & \textbf{97.00} & 82.80 & 82.31 & 93.86 & +1.30\% & 79.71 \\
  SiRA & 88.59 & 53.23 & 87.49 & \textbf{71.63} & 78.06 & 63.46 & 70.80 & 84.82 & 81.63 & 87.36 & \textbf{97.00} & \textbf{83.80} & 82.31 & 93.85 & +1.30\% & 80.29 \\
 \cellcolor{green!10}T-REX & \cellcolor{green!10}88.51 & \cellcolor{green!10}53.76 & \cellcolor{green!10}88.32 & \cellcolor{green!10}71.16 & \cellcolor{green!10}\textbf{80.66} & \cellcolor{green!10}\textbf{68.27} & \cellcolor{green!10}71.50 & \cellcolor{green!10}\textbf{85.31} & \cellcolor{green!10}\textbf{82.65} & \cellcolor{green!10}\textbf{89.89} & \cellcolor{green!10}96.00 & \cellcolor{green!10}83.40 & \cellcolor{green!10}\textbf{82.80} & \cellcolor{green!10}\textbf{94.38} & \cellcolor{green!10}\textcolor{red}{\textbf{+0.76\%}} & \cellcolor{green!10}\textcolor{rgb:red,0.13;green,0.55;blue,0.13}{\textbf{81.19}} \\
 \midrule
 \midrule
\rowcolor{gray!10}
\multicolumn{17}{c}{\textit{\textbf{Mistral 7B}}} \\
\midrule
\midrule
  LoRA & \textbf{90.37} & 59.63 & \textbf{90.61} & 72.41 & 86.03 & 65.38 & 76.30 & 87.83 & 81.93 & \textbf{90.97} & 96.00 & 86.20 & \textbf{84.44} & 95.06 & +1.15\% & 83.08 \\
  MoLoRA & 90.14 & 59.05 & 90.49 & 73.82 & 86.50 & 66.35 & 75.50 & 88.52 & 81.63 & 90.61 & 97.00 & \textbf{87.80} & 83.37 & 95.12 & +1.28\% & 83.28 \\
  SiRA & 90.00 & 58.46 & 89.91 & 74.14 & 86.98 & 60.58 & 75.10 & \textbf{89.23} & 82.86 & 89.89 & 96.00 & 87.40 & 83.62 & 95.07 & +1.28\% & 82.80 \\
\cellcolor{green!10}T-REX & \cellcolor{green!10}90.24 & \cellcolor{green!10}\textbf{62.97} & \cellcolor{green!10}90.76 & \cellcolor{green!10}\textbf{74.76} & \cellcolor{green!10}\textbf{87.06} & \cellcolor{green!10}\textbf{70.19} & \cellcolor{green!10}\textbf{78.30} & \cellcolor{green!10}90.70 & \cellcolor{green!10}\textbf{83.57} & \cellcolor{green!10}90.61 & \cellcolor{green!10}\textbf{99.00} & \cellcolor{green!10}87.60 & \cellcolor{green!10}86.08 & \cellcolor{green!10}\textbf{96.21} & \cellcolor{green!10}\textcolor{red}{\textbf{+0.76\%}} & \cellcolor{green!10}\textcolor{rgb:red,0.13;green,0.55;blue,0.13}{\textbf{84.86}} \\ 
 \midrule
 \midrule
\rowcolor{gray!10}
\multicolumn{17}{c}{\textit{\textbf{Yi 6B}}} \\
\midrule
\midrule
  LoRA & \textbf{89.93} & \textbf{62.77} & \textbf{89.08} & 73.82 & 83.35 & 63.46 & 71.20 & 87.70 & 81.93 & 88.81 & 96.00 & 87.60 & 84.36 & 95.18 & +1.18\% & 82.51 \\
  MoLoRA & 89.83 & 61.66 & 88.96 & \textbf{73.98} & 83.82 & 63.46 & 71.70 & 87.92 & 82.45 & 88.81 & 96.00 & 87.20 & 83.37 & 95.10 & +1.33\% & 82.45 \\
  SiRA & 89.44 & 61.79 & 88.96 & 73.51 & 83.50 & \textbf{68.27} & \textbf{72.10} & 87.92 & \textbf{83.01} & 88.09 & 97.00 & \textbf{88.60} & \textbf{85.42} & 95.21 & +1.33\% & 83.06 \\
 \cellcolor{green!10}T-REX & \cellcolor{green!10}89.47 & \cellcolor{green!10}62.57 & \cellcolor{green!10}88.86 & \cellcolor{green!10}73.75 & \cellcolor{green!10}\textbf{83.87} & \cellcolor{green!10}64.50 & \cellcolor{green!10}\textbf{72.10} & \cellcolor{green!10}\textbf{88.05} & \cellcolor{green!10}82.43 & \cellcolor{green!10}\textbf{88.92} & \cellcolor{green!10}\textbf{98.00} & \cellcolor{green!10}87.20 & \cellcolor{green!10}85.06 & \cellcolor{green!10}\textbf{95.68} & \cellcolor{green!10}\textcolor{red}{\textbf{+0.82\%}} & \cellcolor{green!10}\textcolor{rgb:red,0.13;green,0.55;blue,0.13}{\textbf{83.89}} \\
 \midrule
 \midrule
\rowcolor{gray!10}
\multicolumn{17}{c}{\textit{\textbf{Phi-2 2B}}} \\
\midrule
\midrule
  LoRA & 88.00 & 54.80 & 86.36 & 71.94 & 77.90 & \textbf{61.54} & 64.10 & 84.11 & 81.37 & \textbf{87.36} & 96.00 & 83.80 & 79.52 & 91.69 & +1.67\% & 79.18 \\
  MoLoRA & 88.02 & 54.54 & 86.27 & 72.41 & 78.45 & 59.62 & 63.70 & 83.68 & 81.06 & 86.28 & 96.00 & 84.20 & 80.18 & 91.17 & +1.87\% & 78.97 \\
  SiRA & 87.71 & 54.67 & 86.33 & \textbf{72.88} & 78.14 & 60.58 & 64.90 & 84.39 & 81.32 & 87.00 & \textbf{97.00} & 83.00 & 80.02 & 91.48 & +1.87\% & 79.24 \\
 \cellcolor{green!10}T-REX & \cellcolor{green!10}\textbf{88.31} & \cellcolor{green!10}\textbf{55.83} & \cellcolor{green!10}\textbf{86.51} & \cellcolor{green!10}72.66 & \cellcolor{green!10}\textbf{80.21} & \cellcolor{green!10}60.22 & \cellcolor{green!10}\textbf{67.43} & \cellcolor{green!10}\textbf{84.61} & \cellcolor{green!10}\textbf{81.62} & \cellcolor{green!10}86.37 & \cellcolor{green!10}\textbf{97.00} & \cellcolor{green!10}\textbf{84.80} & \cellcolor{green!10}\textbf{81.41} & \cellcolor{green!10}\textbf{92.35} & \cellcolor{green!10}\textcolor{red}{\textbf{+1.15\%}} & \cellcolor{green!10}\textcolor{rgb:red,0.13;green,0.55;blue,0.13}{\textbf{79.95}} \\ 
 \midrule
 \midrule
\rowcolor{gray!10}
\multicolumn{17}{c}{\textit{\textbf{Gemma 2B}}} \\
\midrule
\midrule
  LoRA & 84.53 & 43.50 & \textbf{85.29} & \textbf{71.16} & 67.56 & \textbf{60.58} & 62.50 & 79.54 & \textbf{76.87} & 85.56 & 83.00 & 75.20 & 73.63 & 87.49 & +1.54\% & 74.03 \\
  MoLoRA & 85.00 & 45.20 & 85.11 & 70.06 & 65.59 & 49.04 & 60.50 & 79.92 & 76.87 & 86.28 & 88.00 & 77.00 & 75.43 & 86.92 & +1.70\% & 73.64 \\
  SiRA & 84.74 & 44.55 & 84.83 & 69.75 & 66.85 & 52.88 & 60.30 & 78.94 & 75.64 & 84.12 & 88.00 & 75.80 & 75.51 & 86.29 & +1.70\% & 73.44 \\
\cellcolor{green!10}T-REX & \cellcolor{green!10}\textbf{85.52} & \cellcolor{green!10}\textbf{45.85} & \cellcolor{green!10}84.89 & \cellcolor{green!10}68.65 & \cellcolor{green!10}\textbf{68.67} & \cellcolor{green!10}57.69 & \cellcolor{green!10}\textbf{63.00} & \cellcolor{green!10}\textbf{80.36} & \cellcolor{green!10}76.82 & \cellcolor{green!10}\textbf{88.09} & \cellcolor{green!10}\textbf{89.00} & \cellcolor{green!10}\textbf{77.20} & \cellcolor{green!10}\textbf{76.74} & \cellcolor{green!10}\textbf{87.85} & \cellcolor{green!10}\textcolor{red}{\textbf{+0.96\%}} & \cellcolor{green!10}\textcolor{rgb:red,0.13;green,0.55;blue,0.13}{\textbf{75.02}} \\
 \midrule
 \midrule
\rowcolor{gray!10}
\multicolumn{17}{c}{\textit{\textbf{Tiny-LLaMA 1B}}} \\
\midrule
\midrule
  LoRA & \textbf{83.04} & 41.80 & 79.94 & 60.34 & 58.72 & 47.12& \textbf{51.60} & 75.41 & \textbf{74.16} & 84.84 & 83.00 & \textbf{70.60} & \textbf{73.63} & \textbf{81.99} & +2.24\% & 69.01 \\
  MoLoRA & 81.70 & \textbf{44.61} & 78.69 & 58.62 & 56.75 & \textbf{52.88} & 49.60 & 75.63 & 72.42 & 83.03 & 83.00 & 67.80 & 72.40 & 80.82 & +2.51\% & 68.42 \\
  SiRA & 81.54 & 42.85 & 79.05 & 57.37 & 53.91 & 43.27 & 47.40 & 74.10 & 72.62 & 81.95 & 80.00 & 65.00 & 71.99 & 79.51 & +2.51\% & 66.47 \\
\cellcolor{green!10}T-REX & \cellcolor{green!10}82.78 & \cellcolor{green!10}43.50 & 
\cellcolor{green!10}\textbf{80.61} & \cellcolor{green!10}\textbf{66.61} & \cellcolor{green!10}\textbf{59.43} & 
\cellcolor{green!10}48.08 & \cellcolor{green!10}51.20 & \cellcolor{green!10}\textbf{75.68} & 
\cellcolor{green!10}74.05 & \cellcolor{green!10}\textbf{85.92} & \cellcolor{green!10}\textbf{87.00} & 
\cellcolor{green!10}68.80 & \cellcolor{green!10}73.55 & \cellcolor{green!10}81.80 & \cellcolor{green!10}\textcolor{red}{\textbf{+1.55\%}} & \cellcolor{green!10}\textcolor{rgb:red,0.13;green,0.55;blue,0.13}{\textbf{69.93}} \\
 \bottomrule
\end{tabular}
}
\vspace{-1.5em}
\end{table*}

\vspace{-0.5em}
\subsection{Quantitative Results}
\label{sec:exp-main}
\paragraph{Performance enhancement.}
We evaluate T-REX across 14 datasets on seven LLM backbones, as shown in Table~\ref{tab:my-table}, without using additional prompt examples to ensure fair comparison. T-REX achieves the highest average score (AVG) across all models, showcasing superior multitask capabilities. Specifically, T-REX demonstrates significant performance improvements on challenging tasks such as WSC and ANLI. For instance, T-REX improves WSC performance by 7.54\% on the Mistral 7B model and achieves a substantial gain of 4.80\% on LLaMA-2 7B, showcasing its ability to handle complex reasoning tasks. On the ANLI task, T-REX outperforms all competing methods across multiple backbones, demonstrating exceptional consistency and robustness, with minimal performance variance across different model scales. Additionally, T-REX achieves competitive or superior results on other datasets, such as COPA and CSQA, further affirming its generalization capability. Notably, T-REX achieves these improvements with fewer additional trainable parameters (\#ATP), as evidenced in multiple settings, e.g., only +0.62\% for LLaMA-2 13B and +0.76\% for Mistral 7B, demonstrating its efficiency and scalability. These results solidify T-REX as a robust and effective MTL solution, excelling in both performance and efficiency across diverse LLM backbones.

\begin{wraptable}{r}{0.4\textwidth}
\vspace{-4mm}
\centering
\footnotesize
\renewcommand{\arraystretch}{0.9} 
\captionsetup{font={footnotesize}}
\caption{\label{tab:ablation-params} Computational overhead with different methods on GeForce RTX 3090.}
\vspace{-2mm}
\setlength\tabcolsep{1pt}
        \resizebox{0.4\columnwidth}{!}{%
  \begin{tabular}{c|c|cccc}
            \toprule
            Method   & Rank & \#ATP \textcolor{red}{\textbf{$\downarrow$}} & FLOPs \textcolor{red}{\textbf{$\downarrow$}} & Speed \textcolor{rgb:red,0.0;green,0.8;blue,0.2}{\textbf{$\uparrow$}} & Acc. \textcolor{rgb:red,0.0;green,0.8;blue,0.2}{\textbf{$\uparrow$}}  \\ 
            \midrule
            \midrule
            \rowcolor{gray!10}
            \multicolumn{6}{c}{\textit{\textbf{Mistral 7B}}} \\
            \midrule
            \midrule
            LoRA   & 32+32     & 83.9M  & 8.13G  & 6.87it/s & 83.13       \\ 
            MoLoRA  & 32+32     & 93.8M  & 9.19G & 6.43it/s & 83.28  \\ 
            \midrule
            \cellcolor{green!10}T-REX & \cellcolor{green!10} 4+8 & \cellcolor{green!10}\textbf{39.9M}& \cellcolor{green!10}\textbf{5.13G}  & \cellcolor{green!10}\textbf{6.92it/s} & \cellcolor{green!10}\textbf{84.86}  \\
            \bottomrule
            \end{tabular}
        }
\vspace{-5mm}
\end{wraptable}
\paragraph{Efficiency analysis.}
\label{sec:effi}
We extend our analysis to evaluate the efficiency of T-REX. To this end, we benchmark T-REX against two established baselines by comparing the number of \#ATP and the computational cost during training, measured in FLOPs and throughput, on the Mistral 7B model. The results, summarized in Table~\ref{tab:ablation-params}, demonstrate a significant reduction in computational overhead during training compared to LoRA and MoLoRA (SiRA shares a similar overhead with MoLoRA). Specifically, T-REX only takes 39.9M additional parameters, an additional 5.13G FLOPs, and a training speed of 6.92 iterations per second on a commercial-grade RTX 3090, while achieving superior performance with accuracy increasing from 83.08\% to 84.86\%. These findings highlight that T-REX not only improves computational efficiency but also optimizes resource usage, delivering better performance with a smaller training footprint.
\vspace{-0.5em}

\begin{figure}[t]
\centering
\includegraphics[width=\linewidth]{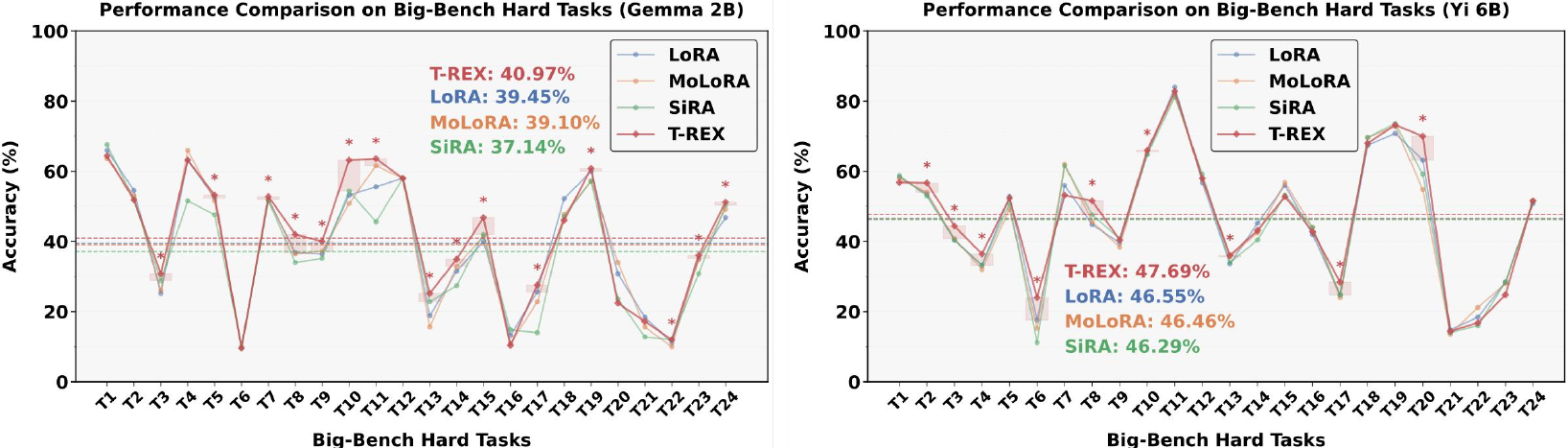}
\vspace{-1.5em}
\caption{Out-of-distribution generalization capabilities of T-REX compared with baselines, including LoRA, MoLoRA, and SiRA, on the BBH dataset based on model backbone (a) Gemma 2B and (b) Yi 6B.}
\label{fig:bbh}
\vspace{-1em}
\end{figure}

\begin{wraptable}{r}{0.4\textwidth}
\vspace{-4mm}
\centering
\footnotesize
\renewcommand{\arraystretch}{0.9} 
\captionsetup{font={footnotesize}}
\caption{\label{tab:ood} OOD and ID data exploration of T-REX based on Tiny-LLaMA 1B.}
\vspace{-2mm}
\setlength\tabcolsep{4pt}
        \resizebox{0.4\columnwidth}{!}{%
             \begin{tabular}{c|c|cccc}
        \toprule
         Method & Data & Drop-A & Drop-B & Drop-C & Drop-D \\ 
         \midrule
         \midrule
        LoRA & \multirow{2}{*}{ID} & 68.58 & 72.43 & 68.45 & 67.15 \\
         \cellcolor{green!10}T-REX &  & \cellcolor{green!10}\textbf{71.22} & \cellcolor{green!10}\textbf{72.61} & \cellcolor{green!10}\textbf{71.69} & \cellcolor{green!10}\textbf{68.03} \\ 
         \midrule
         \midrule
         LoRA & \multirow{2}{*}{OOD} & 43.87 & 52.68 & 50.35 & 59.41 \\
        \cellcolor{green!10}T-REX & &\cellcolor{green!10}\textbf{45.42} & \cellcolor{green!10}\textbf{53.43} & \cellcolor{green!10}\textbf{50.55} & \cellcolor{green!10}\textbf{59.71} \\ \bottomrule
        \end{tabular}
        }
\vspace{-5mm}
\end{wraptable}
\paragraph{Intuition generalizability.}
\label{sec:ood}
We investigate the generalization ability of T-REX, attributed to its intuition-guided design. To isolate the impact of intuition, we exclude the Mix-and-Match from T-REX, focusing solely on the effectiveness of prior knowledge guidance. Specifically, we randomly exclude five sub-datasets from the training set in each experiment, treating the excluded $\textit{Drop-X}$ datasets as out-of-distribution (OOD) and the remaining ones as in-distribution (ID). As shown in Table~\ref{tab:ood}, T-REX consistently achieves improvements in both OOD and ID scenarios. Notably, for the $\textit{Drop-A}$ setting, T-REX improves ID performance by 2.48\% and OOD performance by 1.55\%, respectively. These results highlight the strong generalization capability of our intuition-guided routing.

In this study, we evaluated the OOD generalization capabilities of our proposed T-REX on the Big-Bench Hard (BBH) dataset~\cite{suzgun2022bbh} in a zero-shot setting, without any fine-tuning. The results\footnote{Zero-value points were excluded for clarity in visualization.}, presented in Fig.~\ref{fig:bbh}, demonstrate that T-REX consistently outperforms competing methods across various LLM backbones. This superior performance is attributed to its ability to effectively integrate multitask priors into the MoE, thereby enhancing OOD knowledge acquisition. For instance, T-REX achieves accuracies of 40.97\% and 47.59\% on Gemma 2B and Yi 6B, consistently surpassing LoRA by over 1\%. Detailed results for the figures and tables are shown in Appendix~\ref{sec:supp-drop-out} and ~\ref{sec:supp-bbh}.

\subsection{Ablation Studies}
\label{sec:exp-ablation}

\begin{wraptable}{r}{0.4\textwidth}
\vspace{-5mm}
\centering
\footnotesize
\renewcommand{\arraystretch}{0.9} 
\captionsetup{font={footnotesize}}
\caption{\label{tab:mam} Computational costs with different MaM strategies on GeForce RTX 3090.}
\vspace{-2mm}
\setlength\tabcolsep{2pt}
        \resizebox{0.4\columnwidth}{!}{%
            \begin{tabular}{c|c|cccc}
            \toprule
            Method   & Experts & \#ATP \textcolor{red}{\textbf{$\downarrow$}} & FLOPs \textcolor{red}{\textbf{$\downarrow$}} & Speed \textcolor{rgb:red,0.0;green,0.8;blue,0.2}{\textbf{$\uparrow$}} & Acc. \textcolor{rgb:red,0.0;green,0.8;blue,0.2}{\textbf{$\uparrow$}}  \\ 
            \midrule
            \midrule
            \rowcolor{gray!10}
            \multicolumn{6}{c}{\textit{\textbf{Tiny-LLaMA 1B}}} \\
            \midrule
            \midrule
            \multirow{3}{*}{T-REX}   & 8$\times$8     & 25.23M & 4.53G & 9.54 it/s &  \textbf{70.75}      \\ 
              & 2$\times$16     & 12.62M & 2.26G & 10.14 it/s & 67.52  \\ 
             & \cellcolor{green!10}4$\times$8 & \cellcolor{green!10}\textbf{12.62M}& \cellcolor{green!10}\textbf{2.26G} & \cellcolor{green!10}\textbf{10.32 it/s} & \cellcolor{green!10}69.93  \\
            \bottomrule
            \end{tabular}
        }
\vspace{-5mm}
\end{wraptable}
\paragraph{Mix-and-Match strategy.}
\label{sec:ablation-experts}
We conduct additional experiments to evaluate the effectiveness of the Mix-and-Match strategy, which leverages the flexible combination of Rank-1 experts, as detailed in Table~\ref{tab:mam}. Specifically, we analyze three configurations: 8$\times$8, 2$\times$16, and 4$\times$8. Among these, the 8$\times$8 configuration achieves the highest accuracy of 70.75\%, albeit at the cost of increased computational overhead, with an $\times$2 increase in FLOPs. Interestingly, we observe that the closer the ranks of the low-rank matrices are, the better the model's performance. For instance, the 2$\times$16 configuration achieves only 67.52\% accuracy, despite having almost the same computational overhead as the 4$\times$8 configuration. This can be attributed to the increased rank in the final $AGB$ matrix. To ensure a fair comparison in terms of computational efficiency, we adopt the 4$\times$8 configuration as the default setting throughout our experiments. The complete evaluation results are shown in Appendix~\ref{sec:supp-mam}.
\vspace{-0.5em}

\begin{wraptable}{r}{0.4\textwidth}
\vspace{-5mm}
\centering
\footnotesize
\renewcommand{\arraystretch}{0.9} 
\captionsetup{font={footnotesize}}
\caption{\label{tab:ablation} Ablation study and efficiency improvement of T-REX with different modules.}
\vspace{-2mm}
\setlength\tabcolsep{4pt}
        \resizebox{0.4\columnwidth}{!}{%
        \begin{tabular}{cc|cccc}
\toprule
Intu. & R-1 & LLaMA-2 & Mistral & Phi-2 & T-LLaMA \\ 
\midrule
\midrule
\textcolor{red}{\ding{55}} & \textcolor{red}{\ding{55}}  & 79.90      & 83.08      & 79.18           & 69.01              \\
\textcolor{rgb:red,0.0;green,0.8;blue,0.2}{\ding{51}} & \textcolor{red}{\ding{55}}    & 81.02     &   83.66     & 79.84           & 69.92              \\
\textcolor{red}{\ding{55}} & \textcolor{rgb:red,0.0;green,0.8;blue,0.2}{\ding{51}}      & 80.60    &   83.32     & 79.41           & 69.70              \\
\textcolor{rgb:red,0.0;green,0.8;blue,0.2}{\ding{51}} & \textcolor{rgb:red,0.0;green,0.8;blue,0.2}{\ding{51}}    & 81.11 & 83.84  & \textbf{80.00} & \textbf{70.36}     \\ 
\midrule
\rowcolor{green!10}\multicolumn{2}{c|}{+ MaM} & \textbf{81.19} & \textbf{84.86} & 79.95 & 69.93 \\
\multicolumn{2}{c|}{\#ATP} & -41.5\% & -40.6\% & -38.5\% & -38.3\% \\
\bottomrule
\end{tabular}
        }
\vspace{-5mm}
\end{wraptable}
\paragraph{Module functionality.} To validate the superior performance of T-REX, we evaluated the effectiveness of implicit intuition and Rank-1 experts on various LLM backbones, including LLaMA-2 7B, Mistral 7B, Phi-2 2B, and Tiny-LLaMA 1B. As shown in Table~\ref{tab:ablation}, both components consistently enhance the performance of the LLMs. For instance, incorporating implicit intuition and Rank-1 experts into LLaMA-2 7B yields performance gains of up to 1.12\% and 0.70\%, respectively. These components also complement each other, working synergistically to further elevate T-REX’s capabilities. Notably, their combination results in a significant improvement of 1.35\% on the Tiny-LLaMA, underscoring the effectiveness of our approach.
Moreover, when integrating the Mix-and-Match (MaM) mechanism, T-REX achieves an additional performance boost of 1.02\% while reducing parameter overhead by 40.6\% for Mistral 7B. These results underscore the effectiveness and efficiency of T-REX. The complete results for each tasks are shown in Appendix~\ref{sec:supp-rank1-intuition}.
\vspace{-1em}

\section{Conclusions}
We propose T-REX, a method that enhances LLM adaptability in multi-task learning (MTL) by leveraging semantic-aware implicit intuitions and Mix-and-Match (MaM) Rank-1 experts. Through extra-low-rank experts, T-REX enables a linear combination of low-rank matrices, expanding the vector subspace for improved performance and efficiency with the MaM mechanism. By extracting implicit intuition from semantic diversity via text embeddings, it guides all experts toward a generalizable convergence. Comprehensive experiments confirm T-REX’s superior performance in both in-distribution and out-of-distribution scenarios. The limitations are discussed in Appendix~\ref{ap:limitation}.

\bibliography{neurips_2025}
\bibliographystyle{plainnat}

\clearpage


\appendix
\section*{Appendix for Mixture-of-Rank-One-Experts}
\label{appendix}
This document provides supplementary materials for the main paper. In Appendix~\ref{ap:theory}, we present a detailed derivation of the approximation error bound stated in Theorem 4.2. Additionally, we elaborate on the motivation behind integrating the intuition into our T-REX framework in Appendix~\ref{ap:motivation}. In Appendix~\ref{sec:supp-drop-out}, we discuss how our approach consistently improves over LoRA baselines on both in-distribution and out-of-distribution tasks even if some of the training tasks are dropped. Furthermore, Appendix~\ref{sec:supp-mam} includes a comprehensive evaluation of different Mix-and-Match strategies. In Appendix~\ref{sec:supp-rank1-intuition}, we provide detailed results demonstrating the effectiveness of each component in our work, including the intuition and rank-1 experts. In Appendix~\ref{sec:supp-bbh}, we present the Big-Bench Hard (BBH) results, highlighting the generalization capabilities of the proposed T-REX framework. Last but not least, we discuss the limitation of this work in Appendix~\ref{ap:limitation}

\section{Detailed Derivation of Approximation Error Bound}
\label{ap:theory}
This section presents the detailed derivation for the approximation error bound stated in Theorem 1 of Section~\ref{sec:theory}. Recall that we aim to bound the Frobenius norm of the difference between the T-REX adaptation matrix $\Delta W$ and the target adaptation matrix $\Delta W^\star$:
\begin{equation}
\|\Delta W - \Delta W^\star\|_F^2 \leq \|\Delta \mathbf{W}_\mathrm{out}\|_F^2 + C \cdot \frac{1}{IJ} \|\Delta W^\star\|_F^2 + \mathcal{O}\left(\frac{1}{\tau}\right).
\end{equation}

\subsection*{A.1 Subspace Decomposition of $\Delta W^\star$}

The target adaptation matrix $\Delta W^\star \in \mathbb{R}^{m \times n}$ can be decomposed as:
\begin{equation}
\Delta W^\star = \Delta \mathbf{W}_\mathrm{in} + \Delta \mathbf{W}_\mathrm{out},
\end{equation}
where:
- $\Delta \mathbf{W}_\mathrm{in} \in \mathcal{S}_a \otimes \mathcal{S}_b$ is the component within the subspace spanned by the rank-1 components $\{\mathbf{a}_i \mathbf{b}_j^\top\}$, and
- $\Delta \mathbf{W}_\mathrm{out} \perp \mathcal{S}_a \otimes \mathcal{S}_b$ is the residual component orthogonal to the subspace.

This decomposition directly follows from the projection theorem in linear algebra, ensuring that $\Delta \mathbf{W}_\mathrm{in}$ and $\Delta \mathbf{W}_\mathrm{out}$ are orthogonal:
\begin{equation}
\langle \Delta \mathbf{W}_\mathrm{in}, \Delta \mathbf{W}_\mathrm{out} \rangle = 0.
\end{equation}

The Frobenius norm of the difference $\|\Delta W - \Delta W^\star\|_F^2$ can now be split into two terms:
\begin{equation}
\|\Delta W - \Delta W^\star\|_F^2 = \|\Delta \mathbf{W}_\mathrm{in} - \mathbf{A}\mathbf{G}\mathbf{B}^\top\|_F^2 + \|\Delta \mathbf{W}_\mathrm{out}\|_F^2.
\end{equation}

\subsection*{A.2 Approximation of $\Delta \mathbf{W}_\mathrm{in}$ by T-REX}

T-REX constructs the adaptation matrix $\Delta W$ as:
\begin{equation}
\Delta W = \mathbf{A} \mathbf{G} \mathbf{B}^\top = \sum_{i=1}^N \sum_{j=1}^N G_{ij}(\mathbf{x}) \mathbf{a}_i \mathbf{b}_j^\top,
\end{equation}
where $\mathbf{G}(\mathbf{x}) \in \mathbb{R}^{I \times J}$ is the dynamic routing matrix, and $\mathbf{a}_i \in \mathbb{R}^m$, $\mathbf{b}_j \in \mathbb{R}^n$ are the learned basis vectors.

The term $\mathbf{A} \mathbf{G} \mathbf{B}^\top$ spans the subspace $\mathcal{S}_a \otimes \mathcal{S}_b$, which has dimensionality:
\begin{equation}
\mathrm{dim}(\mathcal{S}_a \otimes \mathcal{S}_b) = IJ.
\end{equation}

Using the projection property, $\Delta \mathbf{W}_\mathrm{in}$ can be approximated within $\mathcal{S}_a \otimes \mathcal{S}_b$ as:
\begin{equation}
\Delta \mathbf{W}_\mathrm{in} = \mathbf{A} \mathbf{G} \mathbf{B}^\top + \Delta \mathbf{W}_\mathrm{res},
\end{equation}
where $\Delta \mathbf{W}_\mathrm{res} \in \mathcal{S}_a \otimes \mathcal{S}_b$ represents the residual error due to the limited expressiveness of the rank-1 combinations.  

The Frobenius norm of the residual error $\Delta \mathbf{W}_\mathrm{res}$ depends on the alignment between $\Delta W^\star$ and the subspace $\mathcal{S}_a \otimes \mathcal{S}_b$:
\begin{equation}
\|\Delta \mathbf{W}_\mathrm{res}\|_F^2 \leq C \cdot \frac{1}{IJ} \|\Delta W^\star\|_F^2,
\end{equation}
where $C$ is a constant that depends on the expressiveness of the subspace.

This term decreases as $IJ$ increases, reflecting the improved subspace coverage of the Mix-and-Match mechanism compared to traditional LoRA, which only spans an $N$-dimensional subspace with LoRA rank $N$.

\subsection*{A.3 Residual Error $\Delta \mathbf{W}_\mathrm{out}$}

The component $\Delta \mathbf{W}_\mathrm{out} \perp \mathcal{S}_a \otimes \mathcal{S}_b$ lies orthogonal to the subspace and cannot be captured by the rank-1 combinations. Its contribution to the total error is:
\begin{equation}
\|\Delta \mathbf{W}_\mathrm{out}\|_F^2 = \|\Delta W^\star - \Delta \mathbf{W}_\mathrm{in}\|_F^2 = \|\Delta W^\star - \mathbf{A} \mathbf{G} \mathbf{B}^\top - \Delta \mathbf{W}_\mathrm{res}\|_F^2.
\end{equation}

Since $\Delta \mathbf{W}_\mathrm{out}$ lies outside $\mathcal{S}_a \otimes \mathcal{S}_b$, its contribution to the total error cannot be reduced by increasing $I\times J$. However, in practice, $\Delta \mathbf{W}_\mathrm{out}$ tends to be small if $\Delta W^\star$ aligns well with the learned subspace, making its contribution negligible.

\subsection*{A.4 Impact of Routing Matrix $\mathbf{G}(\mathbf{x})$}

The routing matrix $\mathbf{G}(\mathbf{x})$ dynamically activates combinations of rank-1 components based on the input $\mathbf{x}$. The sharpness of these activations is controlled by the softmax temperature $\tau$:
\begin{equation}
G_{ij}(\mathbf{x}) = \frac{\exp(z_{ij} / \tau)}{\sum_{i',j'} \exp(z_{i'j'} / \tau)},
\end{equation}
where $z_{ij}$ are the logits for the rank-1 components. A higher temperature $\tau$ results in smoother activations across multiple rank-1 components, whereas a lower $\tau$ sharpens the focus on specific components. The influence of $\tau$ on the approximation error is captured by the term $\mathcal{O}(1/\tau)$, reflecting the trade-off between flexibility and alignment in subspace activation.

\subsection*{A.5 Final Error Bound}

Combining the results from the previous sections, the total approximation error is:
\begin{align}
\|\Delta W - \Delta W^\star\|_F^2 & = \|\Delta \mathbf{W}_\mathrm{in} - \mathbf{A} \mathbf{G} \mathbf{B}^\top\|_F^2 + \|\Delta \mathbf{W}_\mathrm{out}\|_F^2 \\
& \leq \|\Delta \mathbf{W}_\mathrm{res}\|_F^2 + \|\Delta \mathbf{W}_\mathrm{out}\|_F^2 \\
& \leq \|\Delta \mathbf{W}_\mathrm{out}\|_F^2 + C \cdot \frac{1}{IJ} \|\Delta W^\star\|_2^2 + \mathcal{O}\left(\frac{1}{\tau}\right).
\end{align}

Here:
- The first term $\|\Delta \mathbf{W}_\mathrm{out}\|_F^2$ quantifies the residual error outside the subspace $\mathcal{S}_a \otimes \mathcal{S}_b$.
- The second term $\frac{1}{IJ} \|\Delta W^\star\|_2^2$ reflects the approximation error within the subspace, which decreases with increasing $IJ$.
- The third term $\mathcal{O}(1/\tau)$ captures the effect of the routing matrix $\mathbf{G}(\mathbf{x})$ on dynamically activating rank-1 components.

This completes the derivation.

\section{Motivations}
\label{ap:motivation}
Research~\cite{gigerenzer2007gut,harteis2008intuition} has shown that beyond \textit{explicit awareness}, humans leverage \textit{implicit intuition} to make swift decisions and instant reactions in critical situations. Yet, capturing this intuitive process within artificial intelligence is challenging due to its intangible essence. In this paper, we assert that an ideal AI intuition representation hinges on three core criteria: \ding{192} it must capture the nuanced semantic variability of each instance; \ding{193} it should be derivable via an autonomous process; and \ding{194} it must seamlessly integrate into LLMs with low computational overhead.

\begin{figure}[t]
\centering
\includegraphics[width=\textwidth]{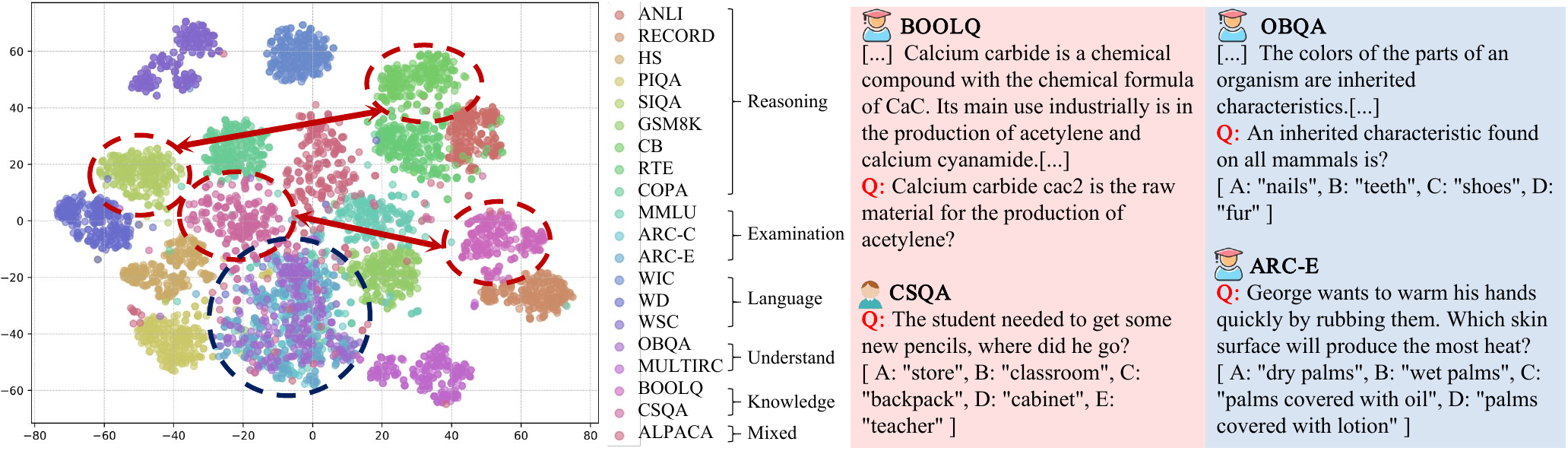}
\caption{Embedding visualization of 20 datasets. Semantic clusters do not align with predefined task groupings.}
\label{fig:motivation}
\end{figure}

In this paper, we integrate \textit{implicit intuition} to improve LLM multitask performance. Initially, we explicitly informed the model with human-derived task categorizations, mapping datasets like BoolQ~\cite{wang2019superglue} to knowledge assessment and ANLI~\cite{nie2019adversarial} to reasoning. However, this method underperformed compared to baseline fine-tuning. This led us to question the effectiveness of traditional task categorizations. 
By visualizing embeddings of 20 datasets across five categories, we found mismatches between expected and actual semantic clusters as in Fig.~\ref{fig:motivation}. Datasets like openbookqa~\cite{OpenBookQA2018} and multirc~\cite{wang2019superglue} for comprehension, and piqa~\cite{Bisk2020} and siqa~\cite{sap2019socialiqa} for reasoning, did not cluster as anticipated. This discrepancy, particularly evident with datasets intended for comprehension and reasoning, highlights the shortcomings of relying on human-annotated task categorizations in guiding multitask learning. Thus, we aim to embed multitask knowledge into the MoE block implicitly via the proposed intuition score mechanism.

\setlength\tabcolsep{2pt}
\begin{table}[b]
\caption{Tiny-LLaMA 1B performance under different training data settings. Gray colors refer to the five datasets excluded during training.}
\label{tab:supp-drop-dataset}
\resizebox{\linewidth}{!}{
\begin{tabular}{c|c|ccccccccccccc|cc}
\toprule
\textbf{Models} & \textbf{Settings} & \textbf{MULTIRC} & \textbf{MMLU} & \textbf{BOOLQ} & \textbf{WIC} & \textbf{WG} & \textbf{WSC} & \textbf{ANLI} & \textbf{PIQA} & \textbf{SIQA} & \textbf{RTE} & \textbf{COPA} & \textbf{OBQA} & \textbf{CSQA} & \textbf{OOD} & \textbf{ID} \\ 
\midrule
\midrule
\multirow{4}{*}{\textbf{LoRA}} & A & \cellcolor[HTML]{C0C0C0}43.36 & \cellcolor[HTML]{C0C0C0}38.41 & 79.82 & 63.64 & 54.14 & 52.88 & \cellcolor[HTML]{C0C0C0}34.70 & 72.69 & 72.47 & 79.42 & 82.00 & 60.20 & \cellcolor[HTML]{C0C0C0}56.67 & 43.87 & 68.58 \\
& B & 81.15 & \cellcolor[HTML]{C0C0C0}38.34 & 80.12 & \cellcolor[HTML]{C0C0C0}53.45 & 53.28 & 63.46 & 51.00 & \cellcolor[HTML]{C0C0C0}67.08 & 72.72 & 81.59 & 86.00 & \cellcolor[HTML]{C0C0C0}47.60 & \cellcolor[HTML]{C0C0C0}56.92 & 52.68 & 72.43 \\
& C & \cellcolor[HTML]{C0C0C0}43.28 & \cellcolor[HTML]{C0C0C0}38.60 & 80.00 & 60.66 & 54.14 & 48.08 & \cellcolor[HTML]{C0C0C0}32.80 & \cellcolor[HTML]{C0C0C0}69.21 & 73.80 & \cellcolor[HTML]{C0C0C0}67.87 & 82.00 & 61.00 & 73.71 & 50.35 & 68.45 \\
& D & 82.51 & 41.80 & 80.80 & 64.42 & 58.80 & 53.85 & 52.70 & \cellcolor[HTML]{C0C0C0}66.38 & \cellcolor[HTML]{C0C0C0}58.50 & 84.48 & 85.00 & \cellcolor[HTML]{C0C0C0}67.40 & \cellcolor[HTML]{C0C0C0}58.72 & 59.41 & 67.15 \\
\midrule
\multirow{4}{*}{\textbf{T-REX}} & A & \cellcolor[HTML]{C0C0C0}42.66 & \cellcolor[HTML]{C0C0C0}39.52 & \cellcolor{green!10}80.52 & \cellcolor{green!10}66.93 & \cellcolor{green!10}56.75 & \cellcolor{green!10}58.65 & \cellcolor[HTML]{C0C0C0}32.50 & \cellcolor{green!10}74.97 & \cellcolor{green!10}73.85 & \cellcolor{green!10}80.87 & \cellcolor{green!10}83.00 & \cellcolor{green!10}65.40 & \cellcolor[HTML]{C0C0C0}57.58 & \cellcolor{green!10}\textbf{45.42} & \cellcolor{green!10}\textbf{71.22} \\
& B & \cellcolor{green!10}82.63 & \cellcolor[HTML]{C0C0C0}38.08 & \cellcolor{green!10}81.35 & \cellcolor[HTML]{C0C0C0}51.57 & \cellcolor{green!10}54.22 & \cellcolor{green!10}57.69 & \cellcolor{green!10}52.20 & \cellcolor[HTML]{C0C0C0}68.34 & \cellcolor{green!10}73.39 & \cellcolor{green!10}83.03 & \cellcolor{green!10}85.00 & \cellcolor[HTML]{C0C0C0}48.40 & \cellcolor[HTML]{C0C0C0}60.77 & \cellcolor{green!10}\textbf{53.43} & \cellcolor{green!10}\textbf{72.61} \\
& C & \cellcolor[HTML]{C0C0C0}46.23 & \cellcolor[HTML]{C0C0C0}39.19 & \cellcolor{green!10}81.07 & \cellcolor{green!10}68.81 & \cellcolor{green!10}57.06 & \cellcolor{green!10}53.85 & \cellcolor[HTML]{C0C0C0}33.20 & \cellcolor[HTML]{C0C0C0}70.62 & \cellcolor{green!10}74.26 & \cellcolor[HTML]{C0C0C0}63.54 & \cellcolor{green!10}85.00 & \cellcolor{green!10}66.20 & \cellcolor{green!10}75.27 & \textbf{50.55} & \textbf{71.69} \\
& D & \cellcolor{green!10}83.56 & \cellcolor{green!10}43.31 & \cellcolor{green!10}82.14 & \cellcolor{green!10}69.28 & \cellcolor{green!10}59.59 & \cellcolor{green!10}48.08 & \cellcolor{green!10}54.50 & \cellcolor[HTML]{C0C0C0}65.18 & \cellcolor[HTML]{C0C0C0}59.26 & \cellcolor{green!10}84.84 & \cellcolor{green!10}87.00 & \cellcolor[HTML]{C0C0C0}70.40 & \cellcolor[HTML]{C0C0C0}58.89 & \cellcolor{green!10}\textbf{59.71} & \cellcolor{green!10}\textbf{68.03} \\ 
\bottomrule
\end{tabular}
}
\end{table}

\section{Dropping out training data}
\label{sec:supp-drop-out}
We randomly drop out five datasets from the training set each time, denoting the dropped datasets as out-of-distribution (OOD) datasets and the remainder as in-distribution (ID). As shown in Table~\ref{tab:supp-drop-dataset}, we demonstrate our method consistently improves average accuracy for both ID and OOD tasks regardless of data configurations on Tiny-LLaMA 1B~\cite{Zhang2024TinyLlamaAO}.
We conducted four independent experiments with different configurations, labeled as A, B, C, and D:
\begin{itemize}
    \item \textbf{Experiment A:} Datasets excluded HellaSwag~\cite{zellers2019hellaswag}, MMLU~\cite{hendryckstest2021}, CommonSenseQA~\cite{talmor-etal-2019-commonsenseqa}, ANLI~\cite{nie2019adversarial}, and MultiRC~\cite{wang2019superglue}, accounting for 31\% of the total training samples.
    \item \textbf{Experiment B:} Datasets excluded WiC~\cite{wang2019superglue}, OpenBookQA~\cite{OpenBookQA2018}, MMLU~\cite{hendryckstest2021}, CommonSenseQA~\cite{talmor-etal-2019-commonsenseqa}, and PiQA~\cite{Bisk2020}, accounting for 37\% of the total training samples.
    \item \textbf{Experiment C:} Datasets excluded PiQA~\cite{Bisk2020}, RTE~\cite{wang2019superglue}, MMLU~\cite{hendryckstest2021}, ANLI~\cite{nie2019adversarial}, and MultiRC~\cite{wang2019superglue}, accounting for 44\% of the total training samples.
    \item \textbf{Experiment D:} Datasets excluded OpenBookQA~\cite{OpenBookQA2018}, SiQA~\cite{Bisk2020}, CommonSenseQA~\cite{talmor-etal-2019-commonsenseqa}, HellaSwag~\cite{zellers2019hellaswag}, and PiQA~\cite{Bisk2020}, accounting for 24\% of the total training samples.
\end{itemize}

\setlength\tabcolsep{3pt}
\begin{table*}[t]
\centering
\caption{Performance evaluation across 14 datasets on Tiny-LLaMA 1B compared with three baselines. MUL. indicates MULTIRC and BOO. indicates BOOLQ.}
\label{tab:supp_mam}
\resizebox{1\textwidth}{!}{%
\begin{tabular}{c|c|cccccccccccccc|c}
\toprule
 Method & Rank & MUL. & MMLU & BOO. & WIC & WG & WSC & ANLI & PIQA & SIQA & RTE & COPA & OBQA & CSQA & HS & AVG \textcolor{rgb:red,0.0;green,0.8;blue,0.2}{\textbf{\textcolor{rgb:red,0.0;green,0.8;blue,0.2}{\textbf{$\uparrow$}}}} \\ 
 \midrule
 \midrule
\rowcolor{gray!10}
\multicolumn{17}{c}{\textit{\textbf{Tiny-LLaMA 1B}}} \\
\midrule
\midrule
  \multirow{3}{*}{T-REX} & 8$\times$8 & \textbf{82.94} & \textbf{43.70}  & \textbf{81.28} & \textbf{68.81} & \textbf{59.83}  & \textbf{55.77} & \textbf{52.80}  & \textbf{77.20}  & \textbf{74.16} & 82.67  & 83.00  & \textbf{73.80} & 72.15 & \textbf{82.43} & \textcolor{rgb:red,0.13;green,0.55;blue,0.13}{\textbf{70.75}} \\
 & 2$\times$16 & 82.57 & 43.24 & 79.91 & 59.72 & 57.70 & 42.31 & 51.20 & 73.83 & 73.34 & 83.75 & 75.00 & 68.60 & 72.81 & 81.30 & 67.52 
 \\
 & \cellcolor{green!10}4$\times$8 & \cellcolor{green!10}82.78 & \cellcolor{green!10}43.50 & 
\cellcolor{green!10}80.61 & \cellcolor{green!10}66.61 & \cellcolor{green!10}59.43 & 
\cellcolor{green!10}48.08 & \cellcolor{green!10}51.20 & \cellcolor{green!10}75.68 & 
\cellcolor{green!10}74.05 & \cellcolor{green!10}\textbf{85.92} & \cellcolor{green!10}\textbf{87.00} & 
\cellcolor{green!10}68.80 & \cellcolor{green!10}\textbf{73.55} & \cellcolor{green!10}81.80 & \cellcolor{green!10}69.93 \\
 \bottomrule
\end{tabular}
}
\end{table*}

\section{Analysis of different Mix-and-Match strategies}
\label{sec:supp-mam}
We report comprehensive results for Tiny-LLaMA-1B using several Mix-and-Match (MaM) configurations on 14 benchmark datasets, as shown in Table~\ref{tab:supp_mam}. The 8$\times$8 setting attains the highest overall accuracy, ranking first on 9 of the 14 tasks. Nevertheless, the 4$\times$8 variant clearly outperforms all other settings on RTE, COPA, and CSQA, where it leads by sizable margins. Although 8$\times$8 is marginally stronger on average, it incurs about 50\% more parameter overhead and correspondingly higher compute cost than 4$\times$8. To achieve a better accuracy-versus-efficiency trade-off, we therefore adopt the 4$\times$8 configuration in all subsequent experiments.

\section{Synergistic Effects of intuition and rank-1 experts}
\label{sec:supp-rank1-intuition}
As elaborated in the main paper, both implicit intuition and rank-1 experts elevate the LLM's performance on average accuracy across 14 datasets. We provide more accurate details on each subset in this section, shown in Table~\ref{tab:supp-ablation-intuition-rank1}.
For our study, we select three LLMs of varying model sizes, ranging from 1 billion to 7 billion parameters. Our results indicate that implicit intuition and rank-1 experts individually boost model performance. Specifically, Intuition is tailored for expert routing, optimizing the model’s decision-making pathways, while Rank-1 Experts facilitate more efficient parameter utilization across the model's architecture. The synergistic combination of these approaches consistently delivers optimal results.

\setlength\tabcolsep{2pt}
\begin{table*}[]
\caption{Comparative analysis of the contributions of rank-1 and intuition across different models: LLaMA-2 7B, Phi-2 2.7B, and Tiny-LLaMA 1B. Rank-1, intuition, and the Mix-and-Match mechanism individually contribute to the final result.}
\label{tab:supp-ablation-intuition-rank1}
\resizebox{\linewidth}{!}{
\begin{tabular}{c|cc|cccccccccccccc|c}
\toprule
\textbf{Model} & \textbf{Intuition} & \textbf{Rank-1} & \textbf{MUL.} & \textbf{MMLU} & \textbf{BOO.} & \textbf{WIC} & \textbf{WG} & \textbf{WSC} & \textbf{ANLI} & \textbf{PIQA} & \textbf{SIQA} & \textbf{RTE} & \textbf{COPA} & \textbf{OBQA} & \textbf{CSQA} & \textbf{HS} & \textbf{AVG} \\ 
\midrule
\midrule
\multirow{5}{*}{\textbf{\begin{tabular}[c]{@{}c@{}}Llama 2\\ 7B\end{tabular}}} & \textcolor{red}{\ding{55}} & \textcolor{red}{\ding{55}} & 88.00 & 52.71 & 88.56 & 70.69 & 80.51 & 57.69 & 73.80 & 84.93 & 81.47 & 86.28 & 95.00 & 82.20 & 83.29 & 93.50 & 79.90 \\
 & \textcolor{rgb:red,0.0;green,0.8;blue,0.2}{\ding{51}} & \textcolor{red}{\ding{55}} & 88.57 & 54.21 & 88.62 & 74.45 & 83.66 & 60.58 & 72.10 & 84.49 & 81.37 & 88.81 & 96.00 & 84.40 & 82.88 & 94.18 & 81.02 \\
 & \textcolor{red}{\ding{55}} & \textcolor{rgb:red,0.0;green,0.8;blue,0.2}{\ding{51}} & 89.21 & 55.32 & 88.29 & 73.04 & 81.69 & 57.69 & 71.70 & 83.68 & 81.37 & 88.09 & 96.00 & 84.80 & 83.05 & 94.41 & 80.60 \\
 & \textcolor{rgb:red,0.0;green,0.8;blue,0.2}{\ding{51}} & \textcolor{rgb:red,0.0;green,0.8;blue,0.2}{\ding{51}} & 89.13 & 54.54 & 88.44 & 73.51 & 82.64 & 58.65 & 72.80 & 85.69 & 81.99 & 89.17 & 97.00 & 84.20 & 83.29 & 94.45 & \textbf{81.11} \\
  & \multicolumn{2}{c|}{\cellcolor{green!10}+ Mix-and-Match} & \cellcolor{green!10}88.51 & \cellcolor{green!10}53.76 & \cellcolor{green!10}88.32 & \cellcolor{green!10}71.16 & \cellcolor{green!10}\textbf{80.66} & \cellcolor{green!10}\textbf{68.27} & \cellcolor{green!10}71.50 & \cellcolor{green!10}\textbf{85.31} & \cellcolor{green!10}\textbf{82.65} & \cellcolor{green!10}\textbf{89.89} & \cellcolor{green!10}96.00 & \cellcolor{green!10}83.40 & \cellcolor{green!10}\textbf{82.80} & \cellcolor{green!10}\textbf{94.38} & \cellcolor{green!10}\textbf{81.19} \\ 
  \midrule
\multirow{5}{*}{\textbf{\begin{tabular}[c]{@{}c@{}}Phi-2\\ 2.7B\end{tabular}}} & \textcolor{red}{\ding{55}} & \textcolor{red}{\ding{55}} & 88.00 & 54.80 & 86.36 & 71.94 & 77.90 & 61.54 & 64.10 & 84.11 & 81.37 & 87.36 & 96.00 & 83.80 & 79.52 & 91.69 & 79.18 \\
 & \textcolor{rgb:red,0.0;green,0.8;blue,0.2}{\ding{51}} & \textcolor{red}{\ding{55}} & 88.20 & 55.06 & 87.09 & 73.04 & 79.87 & 57.69 & 67.80 & 83.90 & 80.81 & 88.09 & 97.00 & 86.40 & 80.51 & 92.29 & 79.84 \\
 & \textcolor{red}{\ding{55}} & \textcolor{rgb:red,0.0;green,0.8;blue,0.2}{\ding{51}} & 88.37 & 56.17 & 87.22 & 72.10 & 80.11 & 54.81 & 66.40 & 84.49 & 81.68 & 86.64 & 95.00 & 85.20 & 80.92 & 92.57 & 79.41 \\
 & \textcolor{rgb:red,0.0;green,0.8;blue,0.2}{\ding{51}} & \textcolor{rgb:red,0.0;green,0.8;blue,0.2}{\ding{51}} & 88.51 & 53.76 & 88.32 & 71.16 & \textbf{80.66} & \textbf{68.27} & 71.50 & \textbf{85.31} & \textbf{82.65} & \textbf{89.89} & 96.00 & 83.40 & \textbf{82.80} & \textbf{94.38} & \textbf{81.19} \\ 
& \multicolumn{2}{c|}{\cellcolor{green!10}+ Mix-and-Match} & \cellcolor{green!10}\textbf{88.31} & \cellcolor{green!10}\textbf{55.83} & \cellcolor{green!10}\textbf{86.51} & \cellcolor{green!10}72.66 & \cellcolor{green!10}\textbf{80.21} & \cellcolor{green!10}60.22 & \cellcolor{green!10}\textbf{67.43} & \cellcolor{green!10}\textbf{84.61} & \cellcolor{green!10}\textbf{81.62} & \cellcolor{green!10}86.37 & \cellcolor{green!10}\textbf{97.00} & \cellcolor{green!10}\textbf{84.80} & \cellcolor{green!10}\textbf{81.41} & \cellcolor{green!10}\textbf{92.35} & \cellcolor{green!10}79.95 \\ 
 \midrule
\multirow{5}{*}{\textbf{\begin{tabular}[c]{@{}c@{}}Tiny-LLaMA\\ 1B\end{tabular}}} & \textcolor{red}{\ding{55}} & \textcolor{red}{\ding{55}} & 83.04 & 41.80 & 79.94 & 60.34 & 58.72 & 47.12 & 51.60 & 75.41 & 74.16 & 84.84 & 83.00 & 70.60 & 73.63 & 81.99 & 69.01 \\
 & \textcolor{rgb:red,0.0;green,0.8;blue,0.2}{\ding{51}} & \textcolor{red}{\ding{55}} & 83.19 & 43.37 & 81.47 & 61.44 & 59.75 & 50.96 & 52.40 & 75.63 & 75.08 & 85.20 & 82.00 & 71.40 & 73.79 & 83.26 & 69.92 \\
 & \textcolor{red}{\ding{55}} & \textcolor{rgb:red,0.0;green,0.8;blue,0.2}{\ding{51}} & 83.27 & 43.63 & 80.43 & 63.79 & 59.35 & 52.88 & 51.30 & 75.63 & 74.10 & 83.39 & 81.00 & 71.00 & 73.05 & 82.97 & 69.70 \\
 & \textcolor{rgb:red,0.0;green,0.8;blue,0.2}{\ding{51}} & \textcolor{rgb:red,0.0;green,0.8;blue,0.2}{\ding{51}} & 83.25 & 42.72 & 80.95 & 64.42 & 60.85 & 49.04 & 52.40 & 75.52 & 74.51 & 85.56 & 88.00 & 71.80 & 73.05 & 82.92 & \textbf{70.36} \\ 
& \multicolumn{2}{c|}{\cellcolor{green!10}+ Mix-and-Match} & \cellcolor{green!10}82.78 & \cellcolor{green!10}43.50 & 
\cellcolor{green!10}\textbf{80.61} & \cellcolor{green!10}\textbf{66.61} & \cellcolor{green!10}\textbf{59.43} & 
\cellcolor{green!10}48.08 & \cellcolor{green!10}51.20 & \cellcolor{green!10}\textbf{75.68} & 
\cellcolor{green!10}74.05 & \cellcolor{green!10}\textbf{85.92} & \cellcolor{green!10}\textbf{87.00} & 
\cellcolor{green!10}68.80 & \cellcolor{green!10}73.55 & \cellcolor{green!10}81.80 & \cellcolor{green!10}69.93 \\ 
\bottomrule
\end{tabular}
}
\end{table*}

\section{Big-Bench Hard results}
\label{sec:supp-bbh}
As elaborated in the main paper, our proposed T-REX demonstrates remarkable out-of-distribution (OOD) generalization capabilities, surpassing other methods, including LoRA~\cite{Hu2021LoRALA}, MoLoRA~\cite{zadouri2023pushing}, and SiRA~\cite{zhu2023sira}, on the Big-Bench Hard (BBH) dataset. Detailed comparative results are presented in Table~\ref{tab:supp-bbh-yi-gemma,tab:supp-bbh-mistral-llama,tab:supp-bbh-tinyllama-phi}.
Our evaluation, conducted on a diverse set of models including Yi 6B~\cite{young2024yi}, Gemma 2B~\cite{Mesnard2024GemmaOM}, Mistral 7B~\cite{Jiang2023Mistral7}, Llama 2 7B~\cite{Touvron2023Llama2O}, Tiny-LLaMA 1B~\cite{Zhang2024TinyLlamaAO}, and Phi-2~\cite{javaheripi2023phi}, is performed without finetuing on BBH samples. We observe superior OOD performance with our T-REX models, which achieves 1.52\% enhancement over competing methods.

\section{Limitations}
\label{ap:limitation}
In this section, we discuss the limitations of our proposed T-REX. First, our experiments are limited to LLMs and do not extend to other modalities, such as large VLMs. However, given the similarities in model structures and training strategies between LLMs and VLMs, we believe that T-REX can be seamlessly extended to VLMs with minimal modifications.  
In addition, the Mix-and-Match strategy increases the number of experts, which in turn enlarges the size of the router. Nevertheless, leveraging the intuition-based design, the linear layer within the router can be implemented using low-rank matrices, effectively reducing parameter overhead without significantly impacting model performance.

\section{Broader impacts}
\label{ap:broader}
Our proposed T-REX framework aims to make LLMs more efficient and adaptable across diverse multitask settings, potentially lowering the entry barrier for deploying state-of-the-art language technologies in a wide range of domains. By reducing computational and parameter overheads, T-REX can foster broader access to high-quality language models, benefitting areas such as education, healthcare, accessibility technologies, and multilingual applications where resource constraints have traditionally been a bottleneck.

Nevertheless, the increased efficiency and scalability enabled by T-REX may also amplify existing societal concerns related to LLMs. Lowering resource requirements could accelerate the proliferation of LLM-powered systems, which, if not carefully monitored, might exacerbate challenges such as the spread of misinformation, automated content generation for malicious purposes, or the magnification of biases present in training data. Furthermore, more flexible and generalizable models may be deployed in sensitive or high-stakes contexts where robustness and transparency are vital.

\setlength\tabcolsep{6pt}
\begin{table*}[h]
\centering
\caption{Zero-shot generalization on the Big-Bench Hard (BBH) benchmark. Using diverse models, we evaluate the generalization capabilities of various methods, including LoRA, MoLoRA, SiRA, and our T-REX. This table demonstrates the results of Yi 6B and Gemma 2B. }
\label{tab:supp-bbh-yi-gemma}
\resizebox{\linewidth}{!}{
\begin{tabular}{c|cccc|cccc}
\toprule
\multirow{2}{*}{\textbf{Big-Bench Hard}} & \multicolumn{4}{c|}{\textbf{Yi 6B}} & \multicolumn{4}{c}{\textbf{Gemma 2B}} \\ \cmidrule(l){2-9} 
 & \textbf{LoRA} & \textbf{MoLoRA} & \textbf{SiRA} & \textbf{T-REX} & \textbf{LoRA} & \textbf{MoLoRA} & \textbf{SiRA} & \textbf{T-REX} \\ \midrule
 \midrule
\textbf{boolean\_expressions} & 58.40 & 58.00 & 58.80 & \cellcolor{green!10}56.80 & 66.00 & 63.60 & 67.60 & \cellcolor{green!10}64.40 \\
\textbf{causal\_judgement} & 53.48 & 54.01 & 52.94 & \cellcolor{green!10}56.68 & 54.55 & 52.94 & 51.87 & \cellcolor{green!10}51.87 \\
\textbf{date\_understanding} & 40.40 & 40.80 & 40.40 & \cellcolor{green!10}44.40 & 25.20 & 26.00 & 28.80 & \cellcolor{green!10}30.80 \\
\textbf{disambiguation\_qa} & 33.20 & 32.00 & 33.20 & \cellcolor{green!10}36.40 & 63.20 & 66.00 & 51.60 & \cellcolor{green!10}63.20 \\
\textbf{dyck\_languages} & 0.00 & 0.00 & 0.00 & \cellcolor{green!10}0.00 & 0.00 & 0.00 & 0.00 & \cellcolor{green!10}0.00 \\
\textbf{formal\_fallacies} & 52.80 & 49.20 & 50.80 & \cellcolor{green!10}52.40 & 52.40 & 51.60 & 47.60 & \cellcolor{green!10}53.20 \\
\textbf{geometric\_shapes} & 17.60 & 15.20 & 11.20 & \cellcolor{green!10}24.00 & 9.60 & 10.00 & 10.00 & \cellcolor{green!10}9.60 \\
\textbf{hyperbaton} & 56.00 & 62.00 & 61.60 & \cellcolor{green!10}53.20 & 52.00 & 52.00 & 51.60 & \cellcolor{green!10}52.80 \\
\textbf{logical\_deduction\_five\_objects} & 44.80 & 45.60 & 47.60 & \cellcolor{green!10}51.60 & 36.80 & 36.40 & 34.00 & \cellcolor{green!10}42.00 \\
\textbf{logical\_deduction\_seven\_objects} & 39.60 & 38.40 & 40.80 & \cellcolor{green!10}40.40 & 36.40 & 37.20 & 35.20 & \cellcolor{green!10}40.00 \\
\textbf{logical\_deduction\_three\_objects} & 64.80 & 65.60 & 64.80 & \cellcolor{green!10}66.00 & 53.20 & 50.80 & 54.40 & \cellcolor{green!10}63.20 \\
\textbf{movie\_recommendation} & 84.00 & 82.00 & 81.20 & \cellcolor{green!10}82.80 & 55.60 & 61.60 & 45.60 & \cellcolor{green!10}63.60 \\
\textbf{multistep\_arithmetic\_two} & 0.00 & 0.00 & 0.00 & \cellcolor{green!10}0.00 & 0.00 & 0.00 & 0.00 & \cellcolor{green!10}0.00 \\
\textbf{navigate} & 56.80 & 57.60 & 59.20 & \cellcolor{green!10}58.00 & 58.00 & 58.00 & 58.00 & \cellcolor{green!10}58.00 \\
\textbf{object\_counting} & 33.60 & 35.60 & 34.00 & \cellcolor{green!10}36.00 & 18.80 & 15.60 & 22.80 & \cellcolor{green!10}25.20 \\
\textbf{penguins\_in\_a\_table} & 45.21 & 42.47 & 40.41 & \cellcolor{green!10}43.15 & 31.51 & 32.88 & 27.40 & \cellcolor{green!10}34.93 \\
\textbf{reasoning\_about\_colored\_objects} & 56.00 & 56.80 & 53.20 & \cellcolor{green!10}52.80 & 40.00 & 41.60 & 42.00 & \cellcolor{green!10}46.80 \\
\textbf{ruin\_names} & 42.00 & 44.00 & 44.00 & \cellcolor{green!10}42.80 & 13.20 & 10.80 & 14.80 & \cellcolor{green!10}10.40 \\
\textbf{salient\_translation\_error\_detection} & 24.80 & 24.00 & 24.80 & \cellcolor{green!10}28.40 & 25.60 & 22.80 & 14.00 & \cellcolor{green!10}27.60 \\
\textbf{snarks} & 67.42 & 69.66 & 69.66 & \cellcolor{green!10}67.98 & 52.25 & 47.75 & 47.19 & \cellcolor{green!10}46.07 \\
\textbf{sports\_understanding} & 70.80 & 72.80 & 73.60 & \cellcolor{green!10}73.20 & 60.00 & 57.20 & 57.20 & \cellcolor{green!10}60.80 \\
\textbf{temporal\_sequences} & 63.20 & 54.80 & 59.20 & \cellcolor{green!10}70.00 & 30.80 & 34.00 & 23.60 & \cellcolor{green!10}22.40 \\
\textbf{tracking\_shuffled\_objects\_five\_objects} & 14.80 & 13.60 & 14.00 & \cellcolor{green!10}14.40 & 18.40 & 15.60 & 12.80 & \cellcolor{green!10}17.20 \\
\textbf{tracking\_shuffled\_objects\_seven\_objects} & 18.40 & 21.20 & 16.00 & \cellcolor{green!10}16.80 & 11.20 & 10.00 & 12.00 & \cellcolor{green!10}12.00 \\
\textbf{tracking\_shuffled\_objects\_three\_objects} & 28.40 & 28.00 & 28.40 & \cellcolor{green!10}24.80 & 35.20 & 34.80 & 30.80 & \cellcolor{green!10}36.00 \\
\textbf{web\_of\_lies} & 50.80 & 51.60 & 51.20 & \cellcolor{green!10}51.60 & 46.80 & 49.20 & 50.40 & \cellcolor{green!10}51.20 \\
\textbf{word\_sorting} & 0.00 & 0.00 & 0.00 & \cellcolor{green!10}0.00 & 0.00 & 0.00 & 0.00 & \cellcolor{green!10}4.40 \\ \midrule
\textbf{AVG} & 41.38 & 41.29 & 41.15 & \cellcolor{green!10}\textbf{42.39} & 35.06 & 34.75 & 33.01 & \cellcolor{green!10}\textbf{36.58} \\ \bottomrule
\end{tabular}
}
\end{table*}

\setlength\tabcolsep{6pt}
\begin{table*}[t]
\caption{Zero-shot generalization on the Big-Bench Hard (BBH) benchmark. Using diverse models, we evaluate the generalization capabilities of various methods, including LoRA, MoLoRA, SiRA, and our T-REX. This table demonstrates the results of Mistral 7B and Llama-2 7B.}
\label{tab:supp-bbh-mistral-llama}
\resizebox{\linewidth}{!}{
\begin{tabular}{c|cccc|cccc}
\toprule
\multirow{2}{*}{\textbf{Big-Bench Hard}} & \multicolumn{4}{c|}{\textbf{Mistral 7B}} & \multicolumn{4}{c}{\textbf{Llama-2 7B}} \\ \cmidrule(l){2-9} 
 & \textbf{LoRA} & \textbf{MoLoRA} & \textbf{SiRA} & \textbf{T-REX} & \textbf{LoRA} & \textbf{MoLoRA} & \textbf{SiRA} & \textbf{T-REX} \\ \midrule
\textbf{boolean\_expressions} & 62.00 & 65.20 & 58.40 & \cellcolor{green!10}63.60 & 58.00 & 58.00 & 56.00 & \cellcolor{green!10}56.80 \\
\textbf{causal\_judgement} & 56.15 & 56.68 & 58.29 & \cellcolor{green!10}56.15 & 57.75 & 56.68 & 56.68 & \cellcolor{green!10}57.22 \\
\textbf{date\_understanding} & 54.00 & 55.60 & 51.60 & \cellcolor{green!10}50.80 & 42.80 & 41.60 & 44.80 & \cellcolor{green!10}40.80 \\
\textbf{disambiguation\_qa} & 36.40 & 52.00 & 44.80 & \cellcolor{green!10}47.60 & 47.60 & 42.80 & 34.40 & \cellcolor{green!10}53.20 \\
\textbf{dyck\_languages} & 3.20 & 1.20 & 2.00 & \cellcolor{green!10}0.80 & 0.00 & 0.00 & 0.00 & \cellcolor{green!10}0.00 \\
\textbf{formal\_fallacies} & 54.80 & 53.60 & 55.20 & \cellcolor{green!10}55.20 & 57.20 & 58.80 & 58.40 & \cellcolor{green!10}58.40 \\
\textbf{geometric\_shapes} & 11.20 & 22.00 & 13.20 & \cellcolor{green!10}14.00 & 34.00 & 37.20 & 38.80 & \cellcolor{green!10}38.00 \\
\textbf{hyperbaton} & 79.20 & 69.20 & 66.80 & \cellcolor{green!10}74.80 & 57.60 & 66.80 & 62.00 & \cellcolor{green!10}60.40 \\
\textbf{logical\_deduction\_five\_objects} & 56.40 & 58.40 & 58.40 & \cellcolor{green!10}62.80 & 54.80 & 52.80 & 52.80 & \cellcolor{green!10}52.80 \\
\textbf{logical\_deduction\_seven\_objects} & 50.40 & 49.20 & 46.80 & \cellcolor{green!10}52.40 & 47.60 & 48.00 & 47.20 & \cellcolor{green!10}46.80 \\
\textbf{logical\_deduction\_three\_objects} & 86.40 & 84.40 & 82.00 & \cellcolor{green!10}86.40 & 70.80 & 72.80 & 75.20 & \cellcolor{green!10}73.20 \\
\textbf{movie\_recommendation} & 60.80 & 60.40 & 59.60 & \cellcolor{green!10}70.40 & 56.80 & 60.40 & 56.40 & \cellcolor{green!10}60.00 \\
\textbf{multistep\_arithmetic\_two} & 0.00 & 0.40 & 0.00 & \cellcolor{green!10}0.00 & 0.00 & 0.00 & 0.40 & \cellcolor{green!10}0.00 \\
\textbf{navigate} & 57.20 & 60.00 & 47.20 & \cellcolor{green!10}58.00 & 46.80 & 42.80 & 57.60 & \cellcolor{green!10}42.80 \\
\textbf{object\_counting} & 38.80 & 33.60 & 40.40 & \cellcolor{green!10}34.00 & 30.00 & 30.80 & 28.80 & \cellcolor{green!10}34.40 \\
\textbf{penguins\_in\_a\_table} & 54.11 & 54.11 & 50.00 & \cellcolor{green!10}54.11 & 41.10 & 40.41 & 43.15 & \cellcolor{green!10}41.78 \\
\textbf{reasoning\_about\_colored\_objects} & 59.60 & 59.20 & 64.00 & \cellcolor{green!10}61.60 & 55.60 & 46.00 & 52.00 & \cellcolor{green!10}54.40 \\
\textbf{ruin\_names} & 40.00 & 40.40 & 41.60 & \cellcolor{green!10}38.80 & 52.00 & 42.80 & 54.80 & \cellcolor{green!10}44.40 \\
\textbf{salient\_translation\_error\_detection} & 48.80 & 46.80 & 45.20 & \cellcolor{green!10}49.20 & 39.60 & 38.00 & 33.20 & \cellcolor{green!10}36.00 \\
\textbf{snarks} & 67.42 & 67.98 & 67.42 & \cellcolor{green!10}67.98 & 52.25 & 52.25 & 59.55 & \cellcolor{green!10}53.93 \\
\textbf{sports\_understanding} & 66.40 & 66.00 & 62.80 & \cellcolor{green!10}66.40 & 60.00 & 58.80 & 58.00 & \cellcolor{green!10}60.00 \\
\textbf{temporal\_sequences} & 43.20 & 50.00 & 50.80 & \cellcolor{green!10}55.60 & 24.00 & 41.20 & 34.80 & \cellcolor{green!10}38.00 \\
\textbf{tracking\_shuffled\_objects\_five\_objects} & 22.00 & 23.20 & 22.00 & \cellcolor{green!10}23.60 & 15.60 & 19.20 & 17.20 & \cellcolor{green!10}22.00 \\
\textbf{tracking\_shuffled\_objects\_seven\_objects} & 21.20 & 17.60 & 19.60 & \cellcolor{green!10}22.00 & 16.00 & 12.80 & 13.60 & \cellcolor{green!10}16.00 \\
\textbf{tracking\_shuffled\_objects\_three\_objects} & 21.60 & 22.40 & 22.40 & \cellcolor{green!10}23.60 & 29.20 & 24.40 & 24.00 & \cellcolor{green!10}24.00 \\
\textbf{web\_of\_lies} & 46.00 & 48.80 & 50.40 & \cellcolor{green!10}47.60 & 49.60 & 50.00 & 49.20 & \cellcolor{green!10}50.40 \\
\textbf{word\_sorting} & 14.80 & 14.40 & 18.40 & \cellcolor{green!10}16.80 & 15.20 & 14.40 & 15.60 & \cellcolor{green!10}18.80 \\ \midrule
\textbf{AVG} & 44.89 & 45.66 & 44.42 & \cellcolor{green!10}\textbf{46.45} & 41.18 & 41.10 & 41.65 & \cellcolor{green!10}\textbf{42.02} \\ \bottomrule
\end{tabular}
}
\end{table*}

\setlength\tabcolsep{6pt}
\begin{table*}[t]
\caption{Zero-shot generalization on the Big-Bench Hard (BBH) benchmark. Using diverse models, we evaluate the generalization capabilities of various methods, including LoRA, MoLoRA, SiRA, and our T-REX. This table demonstrates the results of Tiny-LLaMA and Phi-2. }
\label{tab:supp-bbh-tinyllama-phi}
\resizebox{\linewidth}{!}{
\begin{tabular}{@{}c|cccc|cccc@{}}
\toprule
\multirow{2}{*}{\textbf{Big-Bench Hard}} & \multicolumn{4}{c|}{\textbf{TinyLlama 1B}} & \multicolumn{4}{c}{\textbf{Phi-2}} \\ \cmidrule(l){2-9} 
 & \textbf{LoRA} & \textbf{MoLoRA} & \textbf{SiRA} & \textbf{T-REX} & \textbf{LoRA} & \textbf{MoLoRA} & \textbf{SiRA} & \textbf{T-REX} \\ \midrule
\textbf{boolean\_expressions} & 56.80 & 58.00 & 58.40 & \cellcolor{green!10}55.20 & 77.20 & 78.40 & 79.20 & \cellcolor{green!10}79.60 \\
\textbf{causal\_judgement} & 51.87 & 50.27 & 51.34 & \cellcolor{green!10}53.48 & 58.82 & 56.68 & 56.15 & \cellcolor{green!10}57.75 \\
\textbf{date\_understanding} & 26.00 & 38.40 & 41.20 & \cellcolor{green!10}30.40 & 39.20 & 39.60 & 38.00 & \cellcolor{green!10}40.40 \\
\textbf{disambiguation\_qa} & 30.00 & 30.00 & 31.20 & \cellcolor{green!10}36.80 & 62.80 & 64.80 & 64.80 & \cellcolor{green!10}63.60 \\
\textbf{dyck\_languages} & 0.00 & 0.00 & 0.00 & \cellcolor{green!10}0.00 & 0.00 & 0.00 & 0.00 & \cellcolor{green!10}0.00 \\
\textbf{formal\_fallacies} & 54.00 & 53.20 & 52.40 & \cellcolor{green!10}52.80 & 55.60 & 56.00 & 56.00 & \cellcolor{green!10}55.20 \\
\textbf{geometric\_shapes} & 0.00 & 0.00 & 9.60 & \cellcolor{green!10}0.00 & 23.60 & 27.60 & 29.20 & \cellcolor{green!10}31.60 \\
\textbf{hyperbaton} & 51.60 & 51.60 & 51.60 & \cellcolor{green!10}51.60 & 69.60 & 72.00 & 66.40 & \cellcolor{green!10}68.00 \\
\textbf{logical\_deduction\_five\_objects} & 26.80 & 27.20 & 21.20 & \cellcolor{green!10}28.00 & 48.00 & 50.40 & 54.80 & \cellcolor{green!10}54.00 \\
\textbf{logical\_deduction\_seven\_objects} & 19.20 & 17.20 & 18.40 & \cellcolor{green!10}18.80 & 56.80 & 56.40 & 53.20 & \cellcolor{green!10}51.60 \\
\textbf{logical\_deduction\_three\_objects} & 45.20 & 46.80 & 40.00 & \cellcolor{green!10}47.20 & 76.00 & 73.60 & 73.60 & \cellcolor{green!10}78.80 \\
\textbf{movie\_recommendation} & 55.20 & 54.80 & 56.80 & \cellcolor{green!10}53.20 & 51.20 & 52.80 & 54.80 & \cellcolor{green!10}52.40 \\
\textbf{multistep\_arithmetic\_two} & 0.40 & 0.40 & 0.40 & \cellcolor{green!10}0.00 & 0.40 & 0.80 & 1.20 & \cellcolor{green!10}0.80 \\
\textbf{navigate} & 54.80 & 43.20 & 42.00 & \cellcolor{green!10}56.80 & 46.00 & 43.20 & 46.00 & \cellcolor{green!10}58.00 \\
\textbf{object\_counting} & 5.60 & 7.20 & 5.60 & \cellcolor{green!10}6.40 & 38.80 & 38.00 & 38.40 & \cellcolor{green!10}40.80 \\
\textbf{penguins\_in\_a\_table} & 32.19 & 30.14 & 28.77 & \cellcolor{green!10}32.88 & 48.63 & 50.00 & 45.89 & \cellcolor{green!10}48.63 \\
\textbf{reasoning\_about\_colored\_objects} & 32.40 & 28.80 & 29.20 & \cellcolor{green!10}30.00 & 58.00 & 55.20 & 57.20 & \cellcolor{green!10}56.40 \\
\textbf{ruin\_names} & 6.00 & 6.40 & 5.20 & \cellcolor{green!10}7.60 & 53.60 & 59.20 & 61.20 & \cellcolor{green!10}51.60 \\
\textbf{salient\_translation\_error\_detection} & 10.40 & 11.60 & 11.20 & \cellcolor{green!10}19.20 & 43.60 & 40.80 & 43.60 & \cellcolor{green!10}44.40 \\
\textbf{snarks} & 49.44 & 43.26 & 48.88 & \cellcolor{green!10}51.12 & 66.85 & 68.54 & 73.60 & \cellcolor{green!10}74.16 \\
\textbf{sports\_understanding} & 64.40 & 48.80 & 46.00 & \cellcolor{green!10}63.60 & 56.40 & 58.40 & 56.40 & \cellcolor{green!10}55.20 \\
\textbf{temporal\_sequences} & 9.20 & 8.80 & 10.40 & \cellcolor{green!10}11.60 & 79.60 & 77.20 & 77.20 & \cellcolor{green!10}62.40 \\
\textbf{tracking\_shuffled\_objects\_five\_objects} & 18.00 & 17.20 & 18.40 & \cellcolor{green!10}17.60 & 18.40 & 18.80 & 18.00 & \cellcolor{green!10}20.80 \\
\textbf{tracking\_shuffled\_objects\_seven\_objects} & 14.00 & 12.00 & 13.60 & \cellcolor{green!10}12.80 & 9.60 & 12.00 & 11.60 & \cellcolor{green!10}13.20 \\
\textbf{tracking\_shuffled\_objects\_three\_objects} & 33.60 & 33.20 & 35.20 & \cellcolor{green!10}34.00 & 32.00 & 31.60 & 30.40 & \cellcolor{green!10}31.60 \\
\textbf{web\_of\_lies} & 48.80 & 51.20 & 51.60 & \cellcolor{green!10}50.00 & 52.40 & 53.20 & 51.60 & \cellcolor{green!10}54.00 \\
\textbf{word\_sorting} & 2.00 & 1.20 & 1.60 & \cellcolor{green!10}3.20 & 20.00 & 18.80 & 17.60 & \cellcolor{green!10}18.00 \\ \midrule
\textbf{AVG} & 29.55 & 28.55 & 28.90 & \cellcolor{green!10}\textbf{30.53} & 46.04 & 46.45 & 46.52 & \cellcolor{green!10}\textbf{46.78} \\ \bottomrule
\end{tabular}
}
\end{table*}


\end{document}